\documentclass{article}

\usepackage{microtype}
\usepackage{graphicx}
\usepackage{subcaption}
\usepackage{booktabs} 

\usepackage{hyperref}
\usepackage{multirow}

\usepackage[preprint]{icml2026}

\usepackage{amsmath}
\usepackage{amssymb}
\usepackage{mathtools}
\usepackage{amsthm}

\usepackage[capitalize,noabbrev]{cleveref}

\theoremstyle{plain}

\theoremstyle{definition}

\theoremstyle{remark}

\usepackage[textsize=tiny]{todonotes}
\newcommand{\alg}{\texttt{Workflow-R1}}
\newcommand{\algs}{\texttt{Workflow-R1-Search}}

\usepackage{color}
\usepackage[most]{tcolorbox}
\usepackage{colortbl}
\usepackage{arydshln}

\usepackage{pifont}
\newcommand{\cmark}{\ding{51}}
\newcommand{\xmark}{\ding{55}}

\newtcolorbox[auto counter, number freestyle={\noexpand\arabic{\tcbcounter}}]{mycolorbox}[3][]{%
    fonttitle=\bfseries,
    breakable,
    title=#3,
    #1
}
\icmltitlerunning{Workflow-R1: Group Sub-sequence Policy Optimization for Multi-turn Workflow Construction}

\begin{document}

\twocolumn[
  \icmltitle{Workflow-R1: Group Sub-sequence Policy Optimization for Multi-turn Workflow Construction}



  \icmlsetsymbol{equal}{*}

  \begin{icmlauthorlist}
    \icmlauthor{Mingze Kong}{cuhksz}
    \icmlauthor{Zikun Qu}{cuhksz}
    \icmlauthor{Zhongquan Zhou}{shangjiao}
    \icmlauthor{Pengyu Liang}{tianjin}
    \icmlauthor{Xiang Li}{tianjin}
    \icmlauthor{Zhiwei Shang}{cuhksz}
    \icmlauthor{Zhi Hong}{cuhksz}
    \icmlauthor{Kaiyu Huang}{tongji}
    \icmlauthor{Zhiyong Wang}{edinburgh}
    \icmlauthor{Zhongxiang Dai}{cuhksz}
  \end{icmlauthorlist}

  \icmlaffiliation{cuhksz}{The Chinese University of Hong Kong, Shenzhen}
  \icmlaffiliation{shangjiao}{Shanghai Jiao Tong University}
  \icmlaffiliation{tianjin}{Tianjin University}
  \icmlaffiliation{tongji}{Tongji University}
  \icmlaffiliation{edinburgh}{University of Edinburgh}

    \icmlcorrespondingauthor{Zhongxiang Dai}{daizhongxiang@cuhk.edu.cn}

  \icmlkeywords{Large Language Models, Multi-agent Systems, Reinforcement Learning}

  \vskip 0.3in
]



\printAffiliationsAndNotice{}  

\begin{abstract}
The rapid evolution of agentic workflows has demonstrated strong performance of LLM-based agents in addressing complex reasoning tasks. However, existing workflow optimization methods typically formulate workflow synthesis as a static, one-shot code-centric generation problem. This paradigm imposes excessive constraints on the model's coding capabilities and restricts the flexibility required for dynamic problem-solving. In this paper, we present Workflow-R1, a framework that reformulates workflow construction as a multi-turn, natural language-based sequential decision-making process. To resolve the optimization granularity mismatch inherent in such multi-turn interactions, we introduce Group Sub-sequence Policy Optimization (GSsPO). While explicitly tailored to align with the interleaved Think-Action dynamics of agentic reasoning, GSsPO fundamentally functions as a structure-aware RL algorithm generalizable to a broad class of multi-turn agentic sequential decision-making tasks. By recalibrating the optimization unit to the composite sub-sequence, specifically the atomic Think-Action cycle, it aligns gradient updates with the semantic boundaries of these interactions, ensuring robust learning in complex multi-turn reasoning tasks. Through extensive experiments on multiple QA benchmarks, Workflow-R1 outperforms competitive baselines, validating GSsPO as a generalized solution for sequential reasoning and establishing Workflow-R1 as a promising new paradigm for automated workflow optimization.
\end{abstract}


\section{Introduction}
\label{sec:intro}

Large Language Models (LLMs) have evolved far beyond simple conversational interfaces. By integrating with external tools and structured reasoning paths, they act as autonomous agents capable of solving complex problems \cite{metagpt, wu2024autogen}. The industrial focus has increasingly shifted towards Workflow-based Agents, defined as systems that decompose complex queries into executable graphs of operators such as search, answer, review, and revise. This paradigm has emerged as a robust standard for tackling intricate tasks.

Recognizing this potential, a growing body of research pursues automated workflow optimization, aiming to replace manual prompt engineering with algorithms that autonomously discover optimal interaction patterns. Many recent methods generate a complete workflow program (often as executable code) to orchestrate operators in a single pass, before any operator is executed and its intermediate results are observed \cite{aflow2024, hu2024automated, nie2025weak, zhang2025metaagent, wang2025evoagentx, zhang2025multi, gao2025flowreasoner, ZhangGDesigner, ruimasgpt, wang2025scoreflow, su2025difficulty, zhou2025multi}. While effective for well-defined tasks, this paradigm imposes a fundamental limitation which we term the \textbf{Static Execution Trap}.

In these frameworks, the computational graph is fully determined prior to execution, effectively decoupling the planning phase from the runtime execution. This results in an open-loop system: the agent commits to a complete sequence of operators without access to intermediate observations or operator execution results generated during the process. Consequently, the workflow lacks the adaptability to handle dynamic environments, as the control flow remains rigid regardless of evolving observations. A more principled approach treats workflow construction not as static program synthesis, but as a sequential decision-making process, where the policy is continuously conditioned on the observation trajectory.

\begin{figure*}[t]
    \centering
    \includegraphics[width=\linewidth]{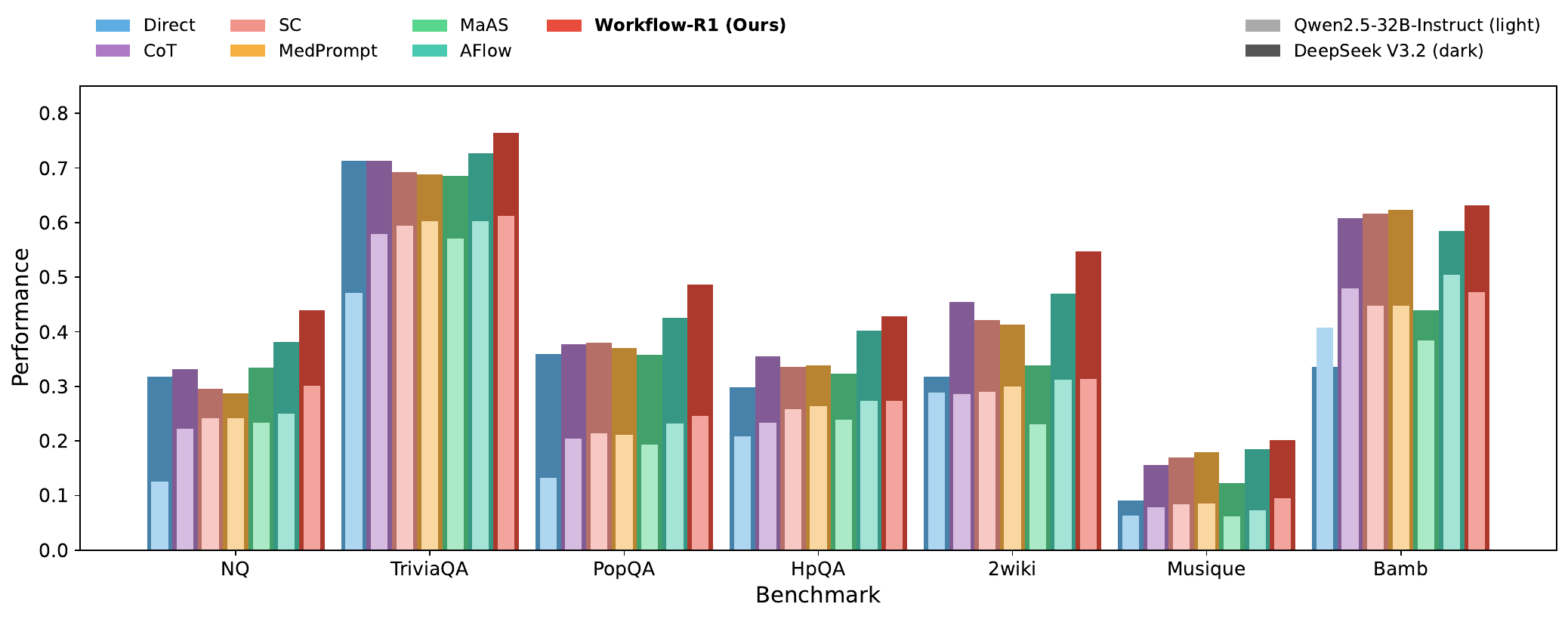}
    \caption{Performance comparison of \alg~against SOTA baselines across seven benchmarks. Lighter and darker shades denote results using Qwen2.5-32B-Instruct and DeepSeek V3.2 as backbones, respectively. \alg demonstrates superior performance across the evaluated benchmarks compared to both standard prompting strategies and advanced workflow optimization methods.}
    \vspace{-5mm}
\end{figure*}

To overcome this inherent rigidity, we draw inspiration from recent advances in iterative reasoning \cite{li2025search, zhang2025router, jin2025search, feng2025retool, zhang2025nemotron} and introduce \alg~. This novel framework facilitates a paradigm shift from static planning to \textbf{Dynamic Interaction}. Specifically, we reframe workflow optimization as a multi-turn conversation between the agent and the environment. Instead of generating a monolithic script, the agent engages in an interleaved cycle of \textit{Thinking, Acting, and Observing}. This closed-loop design ensures that each decision is grounded in operator execution results from preceding steps, empowering the model to dynamically adjust its reasoning path and operator selection in real-time.

Furthermore, in \alg, this interaction is conducted via natural language rather than executable code. This design yields a critical structural advantage: by relieving the agent of the strict constraints of syntactic precision and variable management inherent in code generation, the optimization task becomes significantly more tractable. This abstraction simplifies the decision process, directing the model's focus purely on logical reasoning and operator synergy. Crucially, this simplification enables a cold-start capability: the agent can learn to optimize workflows from scratch via Reinforcement Learning (RL) efficiently, bypassing the need for expert code demonstrations or Supervised Fine-Tuning (SFT) warm-ups.

However, aligning this dynamic, multi-turn decision-making process with reinforcement learning presents a unique challenge: a \textbf{Granularity Mismatch} between standard optimization objectives and the intrinsic semantic structure of agentic reasoning.

Current approaches operate at granularities that create a mismatch with the sequential nature of problem-solving, falling into two problematic extremes. On one end, token-level methods, such as GRPO~\cite{shao2024deepseekmath}, apply optimization at the individual token level. This approach effectively treats complex reasoning chains as loose collections of independent tokens. By enforcing updates on isolated tokens rather than coherent logical units, it risks disrupting the semantic integrity of the think-action process, while overlooking the strong causal dependencies that link a reasoning thought to its subsequent action. On the other end, sequence-level approaches, such as GSPO~\cite{zheng2025group}, treat the entire multi-turn interaction sequence as a monolithic unit. While this aligns the optimization with the final reward signal, it obscures the distinct decision steps inherent in multi-turn workflows. In long-horizon tasks, relying on a single, coarse-grained signal for the whole sequence fails to differentiate effective reasoning steps from erroneous ones, resulting in inefficient learning and slow policy improvement.

We argue that the optimal alignment target is neither the single token nor the holistic sequence, but the Sub-sequence, defined as a decision unit corresponding to a contiguous segment parsed from the full response. In our multi-turn setting, each turn yields a Think-Action pair, which by construction forms a Sub-sequence. To realize this, we propose \textbf{Group Sub-sequence Policy Optimization (GSsPO)}. This structure-aware RL algorithm explicitly aligns the optimization granularity with the agent's decision unit. By treating each Sub-sequence as an atomic optimization unit, GSsPO bridges the gap between fine-grained feedback and global consistency, ensuring that policy updates strictly adhere to the decision boundaries of agentic reasoning.

In summary, this work advances the frontier of automated workflow optimization through three primary contributions:

\textbf{1. A Paradigm Shift to Dynamic Interaction.} We propose \alg, which transforms workflow construction from static program synthesis to a dynamic sequential decision process.

\textbf{2. Structure-Aware Policy Optimization.} We introduce \textbf{Group Sub-sequence Policy Optimization (GSsPO)} to address the granularity mismatch in multi-turn reasoning. Crucially, GSsPO serves as a generalizable algorithmic solution, capable of adapting to and enhancing the sequential reasoning capabilities of diverse multi-turn agent systems.

\textbf{3. Empirical State-of-the-Art.} We demonstrate that \alg~ achieves state-of-the-art performance on complex benchmarks. These results empirically validate the effectiveness of our novel dynamic framework, establishing it as a promising new paradigm for automated workflow optimization.
\vspace{-1mm}
\section{Preliminaries}

In this paper, we formalize the large language model, parameterized by $\theta$, as a policy $\pi_{\theta}$. Given an input prompt $x$ sampled from a dataset $\mathcal{D}$, the model generates a response $y = [y_1, y_2, \dots, y_T]$, where $y_t$ denotes the token at step $t$. The joint probability is factorized as $\pi_{\theta}(y|x) = \prod_{t=1}^T\pi_{\theta}(y_{t}|y_{<t}, x)$.

\subsection{Group Relative Policy Optimization (GRPO)}
GRPO obviates the need for the value model by estimating advantages through group-wise sampling \cite{shao2024deepseekmath}. For each input, it generates multiple outputs and computes relative advantages by normalizing rewards within the group. The optimization objective is defined at the token-level (we omit
the KL regularization for brevity):
\begin{equation}
\begin{aligned}
    &\mathcal{J}_{\text{GRPO}}(\theta) = \mathbb{E}_{x \sim \mathcal{D}, \{y_i\}_{i=1}^G \sim \pi_{\theta_{\text{old}}}(\cdot|x)} \left[ \frac{1}{G} \sum_{i=1}^G \frac{1}{|y_i|} \sum_{t=1}^{|y_i|} \right. \\
    & \left. \min \left( r_{i,t}(\theta) \widehat{A}_{i,t}, \text{clip}(r_{i,t}(\theta), 1{-}\varepsilon, 1{+}\varepsilon) \widehat{A}_{i,t} \right) \vphantom{\sum_{i=1}^G} \right],
\end{aligned}
\end{equation}
\vspace{-5mm}

where the token-level importance sampling ratio $r_{i,t}(\theta)$ is defined as:
\begin{equation}
    r_{i,t}(\theta) = \frac{\pi_{\theta}(y_{i,t} | y_{i,<t}, x)}{\pi_{\theta_{\text{old}}}(y_{i,t} | y_{i,<t}, x)}.
\end{equation}
\vspace{-2mm}

Consistent with prior work, all methods in this paper employ a group-based advantage estimation, where the advantage is computed by standardizing the reward within each sampled group:
\begin{equation}
    \widehat{A}_{i,t} = \frac{r(x, y_i) - \text{mean}\left(\{r(x, y_j)\}_{j=1}^G\right)}{\text{std}\left(\{r(x, y_j)\}_{j=1}^G\right)},
\end{equation}
where $r(x, y_i)$ denotes the reward for the $i$-th sampled response conditioned on input $x$. Consistent with outcome supervision RL, this formulation uniformly propagates the response-level advantage to every token within the generated response.

\subsection{Group Sequence Policy Optimization (GSPO).}
In contrast to GRPO's token-level optimization, GSPO treats the entire response sequence as the optimization unit \cite{zheng2025group}. Accordingly, it applies the importance sampling ratio at the sequence-level, aligning gradient updates with overall generation quality:
\begin{equation}
\begin{aligned}
    &\mathcal{J}_{\text{GSPO}}(\theta) = \mathbb{E}_{x \sim \mathcal{D}, \{y_i\}_{i=1}^G \sim \pi_{\theta_{\text{old}}}(\cdot|x)} \Bigg[ \frac{1}{G} \sum_{i=1}^G \\
    & \min \left( s_i(\theta) \widehat{A}_i, \text{clip} \left( s_i(\theta), 1 - \varepsilon, 1 + \varepsilon \right) \widehat{A}_i \right) \Bigg],
\end{aligned}
\end{equation}
where the sequence-level importance sampling ratio $s_i(\theta)$ is the geometric mean of the per-token likelihood ratios for the full response:
\begin{equation}
\begin{aligned}
    s_i(\theta) &= \left( \frac{\pi_{\theta}(y_i | x)}{\pi_{\theta_{\text{old}}}(y_i | x)} \right)^{\frac{1}{|y_i|}} \\
    &= \exp\left( \frac{1}{|y_i|} \sum_{t=1}^{|y_i|} \log \frac{\pi_{\theta}(y_{i,t} | y_{i,<t}, x)}{\pi_{\theta_{\text{old}}}(y_{i,t} | y_{i,<t}, x)} \right).
\end{aligned}
\end{equation}
The advantage $\widehat{A}_i$ follows the same group-normalized formulation as in GRPO, maintaining consistent reward normalization.
\vspace{-1mm}
\section{Workflow-R1}

In this section, we present the Workflow-R1 framework. We begin with Group Sub-sequence Policy Optimization, our core RL algorithm. We then formalize the multi-turn workflow construction task in the RL setting. Next, we define the reward function for the multi-turn workflow construction. Finally, we specify the prompt template, operator pool and operator interface protocols.

\subsection{GSsPO: Group Sub-sequence Policy Optimization}
\textbf{Motivation.} Existing RL paradigms for LLMs operate at granularities that create a fundamental mismatch with the sequential nature of multi-turn decision-making. Token-level optimization, exemplified by GRPO, treats complex reasoning chains as loose collections of independent tokens. By enforcing updates on isolated tokens rather than coherent decision units, it risks disrupting the semantic integrity of the decision-making process. Conversely, sequence-level optimization, such as GSPO, treats the entire interaction as a decision unit. While this aligns with the final reward, it obscures distinct decision boundaries, failing to differentiate effective decision units from erroneous ones.

To bridge this gap, we posit that optimization granularity must strictly align with the agent's decision boundaries. Therefore, we propose \textbf{Group Sub-sequence Policy Optimization (GSsPO)}. The core design of GSsPO recalibrates importance sampling boundaries to coincide with parsed sub-sequences, which are defined as contiguous segments corresponding to the atomic decision unit. By grounding optimization at this level, GSsPO ensures that gradient updates respect the logical coherence of each decision unit. Furthermore, to mitigate verbosity bias, we explicitly neutralize the confounding factors of action cardinality. This ensures that the optimization process is driven by the quality of the decision unit rather than its length, significantly enhancing model performance in complex multi-turn workflows.

\textbf{Objective Function.}
Our proposed method, GSsPO, recalibrates the optimization granularity to sub-sequences to enhance policy performance. Rather than aggregating gradients at the token or sequence level, GSsPO treats each constituent sub-sequence as a discrete atomic optimization unit. Formally, for a sampled response $y_i$, let $\mathcal{S}_i$ denote the set of sub-sequences parsed from response $i$. The optimization objective function is defined as follows:
\begin{equation}
\begin{aligned}
    &\mathcal{J}_{\text{GSsPO}}(\theta) = \mathbb{E}_{x \sim \mathcal{D}, \{y_i\}_{i=1}^G \sim \pi_{\theta_{\text{old}}}(\cdot|x)}  \Bigg[\frac{1}{G} \sum_{i=1}^G \frac{1}{|\mathcal{S}_i|} \sum_{s \in \mathcal{S}_i} \\
    &\min \left( r_s(\theta) \widehat{A}_s, \text{clip}(r_s(\theta), 1-\varepsilon, 1+\varepsilon) \widehat{A}_s \right) \Bigg],
\end{aligned}
\label{eq:gsspo_objective}
\end{equation}
\vspace{-6mm}

where the inner summation iterates over all sub-sequences $s$ parsed from the sampled sequences. The advantage term $\widehat{A}_s$ inherits the sequence-level outcome, meaning all sub-sequences within the same response share the identical group-normalized advantage value $\widehat{A}_i$. The sub-sequence-level importance sampling ratio $r_s(\theta)$ is formulated as the geometric mean of the token probability ratios:
\begin{equation}
\begin{aligned}
    r_s(\theta) &= \left(\frac{\prod_{t \in s} \pi_\theta(y_t | y_{<t}, x)}{\prod_{t \in s} \pi_{\theta_{\text{old}}}(y_t | y_{<t}, x)}\right)^{\frac{1}{|s|}} \\
    &= \exp\left(\frac{1}{|s|}\sum_{t \in s}\log\frac{\pi_{\theta}(y_t|y_{<t}, x)}{\pi_{\theta_{\text{old}}}(y_t|y_{<t}, x)}\right).
\end{aligned}
\end{equation}
\textbf{Policy Gradient.} To optimize the policy parameters $\theta$, we compute the gradient of the optimization objective $\mathcal{J}_{\text{GSsPO}}$. The policy gradient is derived as follows (clipping is omitted for brevity):
\begin{equation}
\begin{aligned}
    &\nabla_\theta \mathcal{J}_{\text{GSsPO}}(\theta) 
    = \nabla_\theta \mathbb{E}  \Bigg[ \frac{1}{G} \sum_{i=1}^G \frac{1}{|\mathcal{S}_i|} \sum_{s \in \mathcal{S}_i} r_s(\theta) \widehat{A}_s \Bigg] \\
    &= \mathbb{E}  \left[ \frac{1}{G} \sum_{i=1}^G \frac{1}{|\mathcal{S}_i|} \sum_{s \in \mathcal{S}_i} \widehat{A}_s \cdot \nabla_\theta r_s(\theta) \right] \\
    &= \mathbb{E}  \Bigg[ \frac{1}{G} \sum_{i=1}^G \frac{1}{|\mathcal{S}_i|} \sum_{s \in \mathcal{S}_i} r_s(\theta) \widehat{A}_s \cdot \nabla_\theta \log r_s(\theta) \Bigg] \\
    &= \mathbb{E}  \Bigg[ \frac{1}{G} \sum_{i=1}^G \frac{1}{|\mathcal{S}_i|} \sum_{s \in \mathcal{S}_i} \underbrace{\left( \frac{\pi_{\theta}(s | x)}{\pi_{\theta_{\text{old}}}(s | x)} \right)^{\frac{1}{|s|}}}_{r_s(\theta)} \widehat{A}_s \\
    &\quad \cdot \underbrace{\frac{1}{|s|} \sum_{t \in s} \nabla_\theta \log \pi_\theta(y_t | y_{<t}, x)}_{\text{Mean Token Gradient in Sub-sequence}} \Bigg].
\end{aligned}
\label{eq:gsspo_gradient}
\end{equation}

For comparative analysis, the gradient derivations for GRPO and GSPO are provided in appendix \ref{Appx:policy_gradient}. Distinct from existing RL methods, our formulation introduces two fundamental shifts in optimization mechanics. First, regarding importance sampling, GSsPO assigns a unified weight $r_s(\theta)$ to all tokens within the same sub-sequence. Second, the gradient aggregation follows a hierarchical two-stage averaging mechanism: it normalizes the token length within each sub-sequence and subsequently averages by the total sub-sequence count $|\mathcal{S}_i|$.

\begin{figure*}[t]
    \centering
    \includegraphics[width=\textwidth]{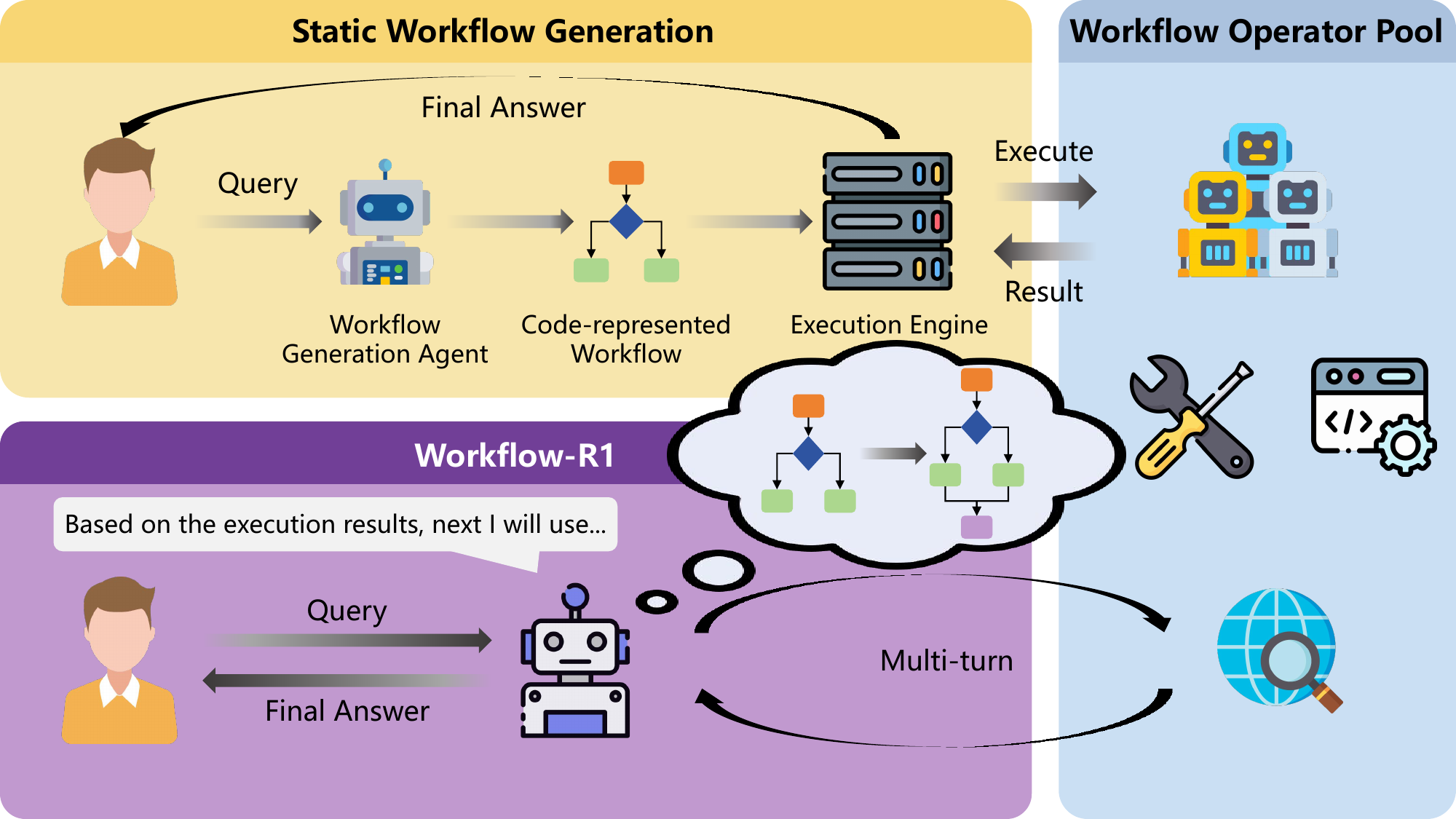} 
    \caption{The top panel shows conventional static workflow generation, where the agent produces a complete executable code-represented workflow. The bottom panel depicts \alg~multi-turn interaction, in which the agent incrementally constructs and adapts the workflow through think-action-observation cycles conditioned on execution results.}
    \label{fig:workflow_r1}
\vspace{-5mm}
\end{figure*}

\subsection{Multi-turn Think-Action Workflow Construction}
\label{subsec:multi_turn_construction}
Distinct from static paradigms that synthesize fixed execution graphs, \alg~reformulates workflow construction as a sequential decision process. The agent autonomously navigates the workflow topology through a recursive cognitive cycle to dynamically discover optimal operator combinations, ensuring the generated topology aligns with specific query requirements.
The interaction is governed by a natural language interface protocol. Upon receiving a query, the agent enters a reasoning phase enclosed within \textcolor{orange}{\texttt{<think>}} tags to formulate a strategic decision: either to conclude via \textcolor{red}{\texttt{<answer>}} or to invoke an operator via \textcolor{blue}{\texttt{<tool>}}. We formally define this \textbf{Think-Action pair} generated within each turn as the \textbf{sub-sequence} in our task. Since the distinct reasoning process and the subsequent operator call jointly form a unified coherent decision unit, we establish this entity as the atomic unit of optimization, where segmentation boundaries are unambiguous (one turn constitutes exactly one sub-sequence).
The abstract Operator Execution Engine intercepts commands and returns operator execution results wrapped in \textcolor{teal}{\texttt{<info>}} tags. This feedback closes the perception-action loop, triggering the subsequent reasoning phase. By eliminating rigid structural boundaries, this architecture empowers the model to aggressively explore operator synergies and adaptively orchestrate resources for deep compositional reasoning.

\begin{table*}[t]
\centering
\resizebox{\textwidth}{!}{%
\begin{tabular}{l c ccc cccc c}
\toprule
\multirow{2}{*}{\textbf{Methods}} & \multirow{2}{*}{\textbf{Multi-turn}} & \multicolumn{3}{c}{\textbf{General QA}} & \multicolumn{4}{c}{\textbf{Multi-Hop QA}} & \multirow{2}{*}{\textbf{Avg.}} \\
\cmidrule(lr){3-5} \cmidrule(lr){6-9}
 & & NQ$^\dagger$ & TriviaQA & PopQA & HpQA$^\dagger$ & 2wiki & Musique & Bamb & \\
\midrule

\multicolumn{9}{l}{\textit{\textbf{Qwen2.5-32B-Instruct}}} \\
\quad Direct & \xmark & 0.126 & 0.471 & 0.133 & 0.208 & 0.289 & 0.063 & 0.408 & 0.243 \\
\quad CoT & \xmark & 0.222 & 0.579 & 0.205 & 0.234 & 0.286 & 0.079 & 0.480 & 0.298 \\
\quad SC (CoT$\times$5) & \xmark & 0.241 & 0.595 & 0.214 & 0.258 & 0.290 & 0.084 & 0.448 & 0.304 \\
\quad MedPrompt & \xmark & 0.242 & 0.603 & 0.211 & 0.264 & 0.300 & 0.086 & 0.448 & 0.308 \\

\addlinespace[0.5ex] \hdashline \addlinespace[0.5ex]

\multicolumn{9}{l}{\textit{\textbf{Agentic Workflow}}} \\
\quad MaAS & \xmark & 0.233 & 0.571 & 0.193 & 0.239 & 0.231 & 0.062 & 0.384 & 0.273 \\
\quad AFlow & \xmark & 0.250 & 0.603 & 0.232 & 0.273 & 0.312 & 0.073 & \textbf{0.504} & 0.321 \\
\rowcolor{gray!15} \quad \textbf{Workflow-R1} & \cmark & \textbf{0.301} & \textbf{0.612} & \textbf{0.246} & \textbf{0.274} & \textbf{0.314} & \textbf{0.095} & 0.472 & \textbf{0.331} \\

\addlinespace[0.5ex] \hdashline \addlinespace[0.5ex]

\multicolumn{9}{l}{\textit{\textbf{Search-Augmented}}} \\
\quad Search-R1 (GRPO)   & \cmark & 0.429 & 0.623 & 0.427 & 0.386 & 0.346 & 0.162 & 0.400 & 0.396 \\
\quad Search-R1 (PPO)    & \cmark & 0.393 & 0.610 & 0.397 & 0.370 & 0.414 & 0.146 & 0.368 & 0.385 \\
\rowcolor{gray!15} \quad \textbf{Workflow-R1-Search} & \cmark & \textbf{0.435} & \textbf{0.733} & \textbf{0.493} & \textbf{0.461} & \textbf{0.633} & \textbf{0.215} & \textbf{0.576} & \textbf{0.507} \\

\bottomrule
\end{tabular}%
}
\caption{Performance comparison of baseline methods on benchmarks. The \textbf{Multi-turn} column indicates whether the method involves iterative interaction. $^\dagger$ represents in-domain datasets.}
\label{tab:baseline_results}
\vspace{-6mm}
\end{table*}

To enable autonomous workflow construction, we employ a reinforcement learning framework. The general optimization objective over the workflow operator pool $\mathcal{W}$ is defined as:
\begin{equation}
\label{eq:optimization_objective}
\begin{aligned}
& \max_{\pi_\theta} \mathbb{E}_{x \sim \mathcal{D}, y \sim \pi_\theta(\cdot|x;\mathcal{W})} \left[ r_\phi(x, y) - \beta \log\frac{\pi_{\theta}(y|x;\mathcal{W})}{\pi_{\text{ref}}(y|x;\mathcal{W})}\right].
\end{aligned}
\end{equation}

This formulation is generic and compatible with various reinforcement learning paradigms, such as GRPO and GSPO. In this work, we adopt our proposed GSsPO as the default algorithm to optimize the workflow construction agent's policy.

\subsection{Reward Design}
We employ a rule-based composite reward function to jointly optimize structural validity and response accuracy. To enable the autonomous exploration of complex workflows while ensuring correctness of format, we enforce strict outcome supervision and formatting constraints. 

\textbf{Format Rewards.}
To enforce adherence to the interface protocol, we apply graded penalties $\mathcal{R}_{\text{Format}}$ based on the severity of syntactic violations:
\begin{equation}
    \mathcal{R}_{\text{Format}} = 
    \begin{cases} 
        0, & \text{if the format is completely correct} \\
        -0.5, & \text{if the tag order is incorrect} \\
        -1, & \text{if the $<$answer$>$ tag is missing}
    \end{cases}
\end{equation}

\textbf{Outcome Supervision.}
We adopt \textbf{Exact Match (EM)} as the outcome reward to verify the correctness of the final answer against the ground truth:
\begin{equation}
    \mathcal{R}_{\text{Outcome}}(y, g) = \text{EM}(y, g).
\end{equation}

Consequently, the final reward signal is constructed by integrating the format constraints with the outcome supervision.  The total reward $r(x, y)$ is formulated as follows:

\begin{equation}
    r_{\phi}(x, y) = \mathcal{R}_{\text{Format}} + \mathcal{R}_{\text{Outcome}}.
\end{equation}
This integration ensures that the model is guided to construct workflows within strict formatting constraints, while retaining the freedom to explore diverse operator combinations. 

\subsection{Operators and Prompt Template}
\label{sec:operators_and_prompts}
We construct the workflow operator pool using a set of operators implemented via natural language interface definitions. This design enables the model to comprehend and execute operator calls directly through natural language instructions. Comprehensive specifications, including detailed operator usages, interface protocols, and the complete training prompt template, are provided in Appendix~\ref{Appx:prompt_template}.
\vspace{-1mm}
\section{Experiments}

\subsection{Datasets}

We evaluate our \alg~and the search-augmented version \algs~on a comprehensive suite of seven datasets designed to test the limits of agentic reasoning. The datasets are classified into two categories: (1) \textbf{General Question Answering}, including \textbf{NQ} \cite{kwiatkowski2019natural}, \textbf{TriviaQA} \cite{joshi2017triviaqa}, and \textbf{PopQA} \cite{mallen2023not}; and (2) \textbf{Multi-Hop Question Answering}, comprising \textbf{HotpotQA} \cite{yang2018hotpotqa}, \textbf{2WikiMultiHopQA} \cite{ho2020constructing}, \textbf{Musique} \cite{trivedi2022musique}, and \textbf{Bamboogle} \cite{press2023measuring}.

\textbf{Data Splits.}
For the training phase, we construct a dataset by sampling a subset from the training splits of NQ and HotpotQA. As detailed in Table \ref{tab:dataset_stats}, this results in a total of 10,000 training instances (4,618 from NQ and 5,382 from HotpotQA). This specific mixture is strategically designed to expose the agent to both General QA and complex multi-hop QA during the RL exploration phase.

\textbf{Metric.} We employ the test sets across all seven benchmarks. This comprehensive protocol allows us to assess both in-domain performance and out-of-domain generalization. We report the Exact Match (EM) score as our metric to quantify correctness.

\begin{figure*}[t]
    \centering
    
    \captionof{table}{Performance comparison of methods on QA benchmarks employing \textbf{DeepSeek V3.2} as the execution model. This ablation evaluates the robustness of workflow optimization across different backbones. $^\dagger$ represents in-domain datasets.}
    \label{tab:baseline_results_deepseek}
    \resizebox{\textwidth}{!}{%
        \begin{tabular}{l ccc cccc c}
        \toprule
        \multirow{2}{*}{\textbf{Methods}} & \multicolumn{3}{c}{\textbf{General QA}} & \multicolumn{4}{c}{\textbf{Multi-Hop QA}} & \multirow{2}{*}{\textbf{Avg.}} \\
        \cmidrule(lr){2-4} \cmidrule(lr){5-8}
         & NQ$^\dagger$ & TriviaQA & PopQA & HpQA$^\dagger$ & 2wiki & Musique & Bamb & \\
        \midrule
        
        \multicolumn{9}{l}{\textit{\textbf{DeepSeek V3.2}}} \\
        \quad Direct      & 0.318 & 0.713 & 0.359 & 0.299 & 0.318 & 0.091 & 0.336 & 0.347 \\
        \quad CoT         & 0.332 & 0.714 & 0.377 & 0.355 & 0.455 & 0.156 & 0.608 & 0.428 \\
        \quad SC (CoT$\times$5) & 0.295 & 0.693 & 0.380 & 0.336 & 0.421 & 0.170 & 0.616 & 0.416 \\
        \quad MedPrompt   & 0.288 & 0.688 & 0.370 & 0.338 & 0.413 & 0.179 & 0.624 & 0.414 \\
        
        \addlinespace[0.5ex] \hdashline \addlinespace[0.5ex]
        
        \multicolumn{9}{l}{\textit{\textbf{Agentic Workflow}}} \\
        \quad MaAS        & 0.335 & 0.685 & 0.358 & 0.323 & 0.339 & 0.123 & 0.440 & 0.372 \\
        \quad AFlow       & 0.382 & 0.727 & 0.426 & 0.402 & 0.470 & 0.185 & 0.584 & 0.454 \\
        \rowcolor{gray!15} \quad \textbf{Workflow-R1 (Ours)} & 0.440 & \textbf{0.765} & 0.487 & 0.428 & 0.548 & 0.202 & \textbf{0.632} & 0.500 \\
        
        \addlinespace[0.5ex] \hdashline \addlinespace[0.5ex]
        
        \multicolumn{9}{l}{\textit{\textbf{Search-Augmented}}} \\
        \quad Search-R1 (GRPO)   & 0.429 & 0.623 & 0.427 & 0.386 & 0.346 & 0.162 & 0.400 & 0.396 \\
        \quad Search-R1 (PPO)    & 0.393 & 0.610 & 0.397 & 0.370 & 0.414 & 0.146 & 0.368 & 0.385 \\
        \rowcolor{gray!15} \quad \textbf{Workflow-R1-Search (Ours)} & \textbf{0.456} & 0.749 & \textbf{0.498} & \textbf{0.460} & \textbf{0.621} & \textbf{0.211} & 0.584 & \textbf{0.511} \\
        \bottomrule
        \end{tabular}%
    }

    \vspace{1em}

    \begin{minipage}[t]{0.48\textwidth}
        \vspace{0pt} 
        \centering
        \resizebox{\linewidth}{!}{%
            \begin{tabular}{lcccc}
                \toprule
                \textbf{Method} & \textbf{NQ$^\dagger$} & \textbf{TriviaQA} & \textbf{HpQA$^\dagger$} & \textbf{Musique} \\
                \midrule
                
                \multicolumn{5}{l}{\textit{Workflow-R1}} \\ 
                GRPO & 0.265 & 0.600 & 0.269 & 0.083 \\
                GSPO & 0.283 & 0.609 & 0.272 & 0.091 \\
                GSsPO & \textbf{0.301} & \textbf{0.612} & \textbf{0.274} & \textbf{0.095} \\
                
                \addlinespace[0.3em] 
                \hdashline 
                \addlinespace[0.3em]
                
                \multicolumn{5}{l}{\textit{Workflow-R1-Search}} \\ 
                GRPO & 0.430 & 0.730 & 0.456 & 0.205 \\
                GSPO & \textbf{0.457} & 0.731 & 0.450 & 0.204 \\
                GSsPO & 0.435 & \textbf{0.733} & \textbf{0.461} & \textbf{0.215} \\
                
                \bottomrule
            \end{tabular}%
        }
        \captionof{table}{Performance comparison of different RL methods across datasets.} 
        \label{tab:workflow_performance}
    \end{minipage}%
    \hfill
    \begin{minipage}[t]{0.48\textwidth}
        \vspace{0pt} 
        \centering
        \includegraphics[width=\linewidth]{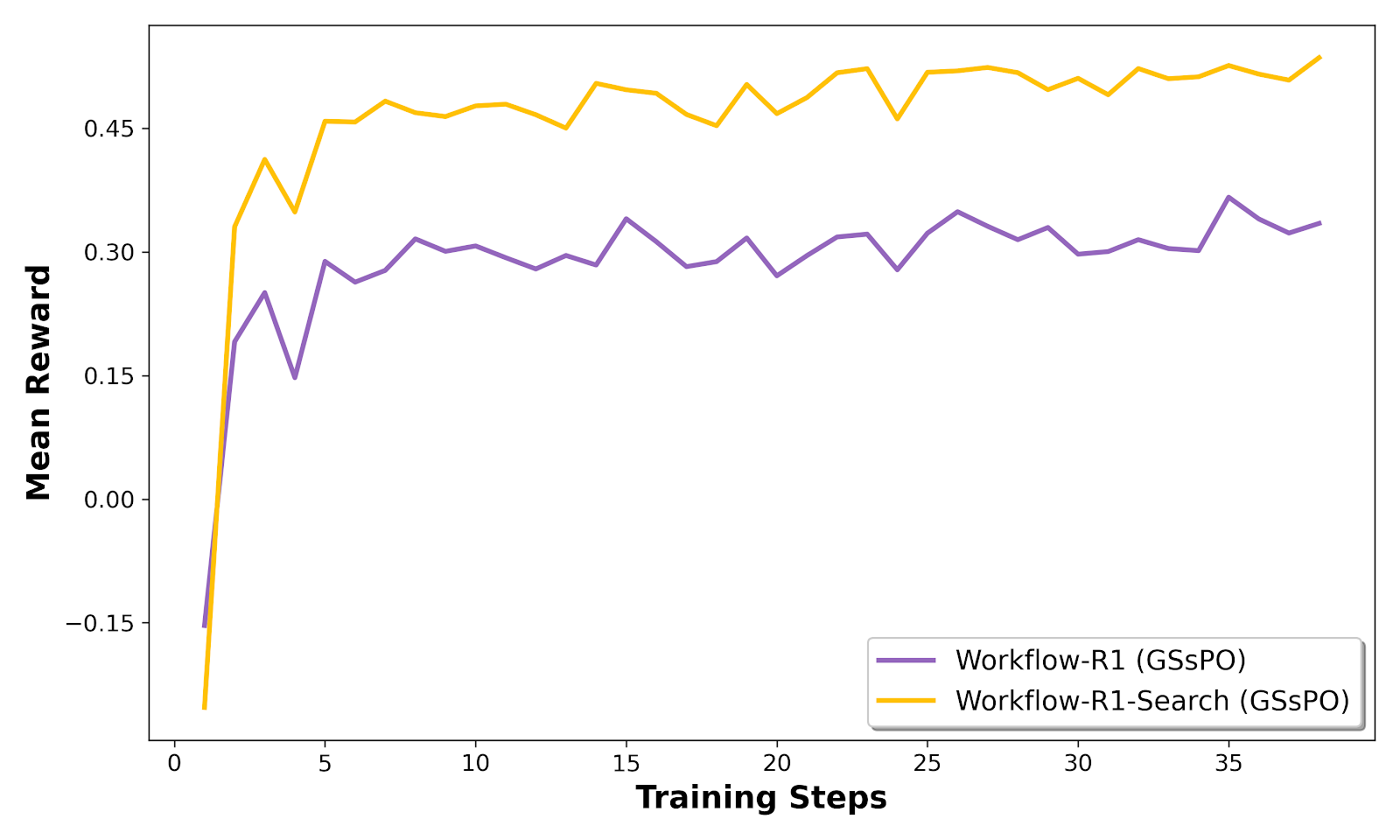}
        \captionof{figure}{Mean reward convergence of GSsPO.}
        \label{fig:convergence_comparison}
    \end{minipage}

\vspace{-6mm}
\end{figure*}

\begin{figure*}
    \centering
    \includegraphics[width=\linewidth]{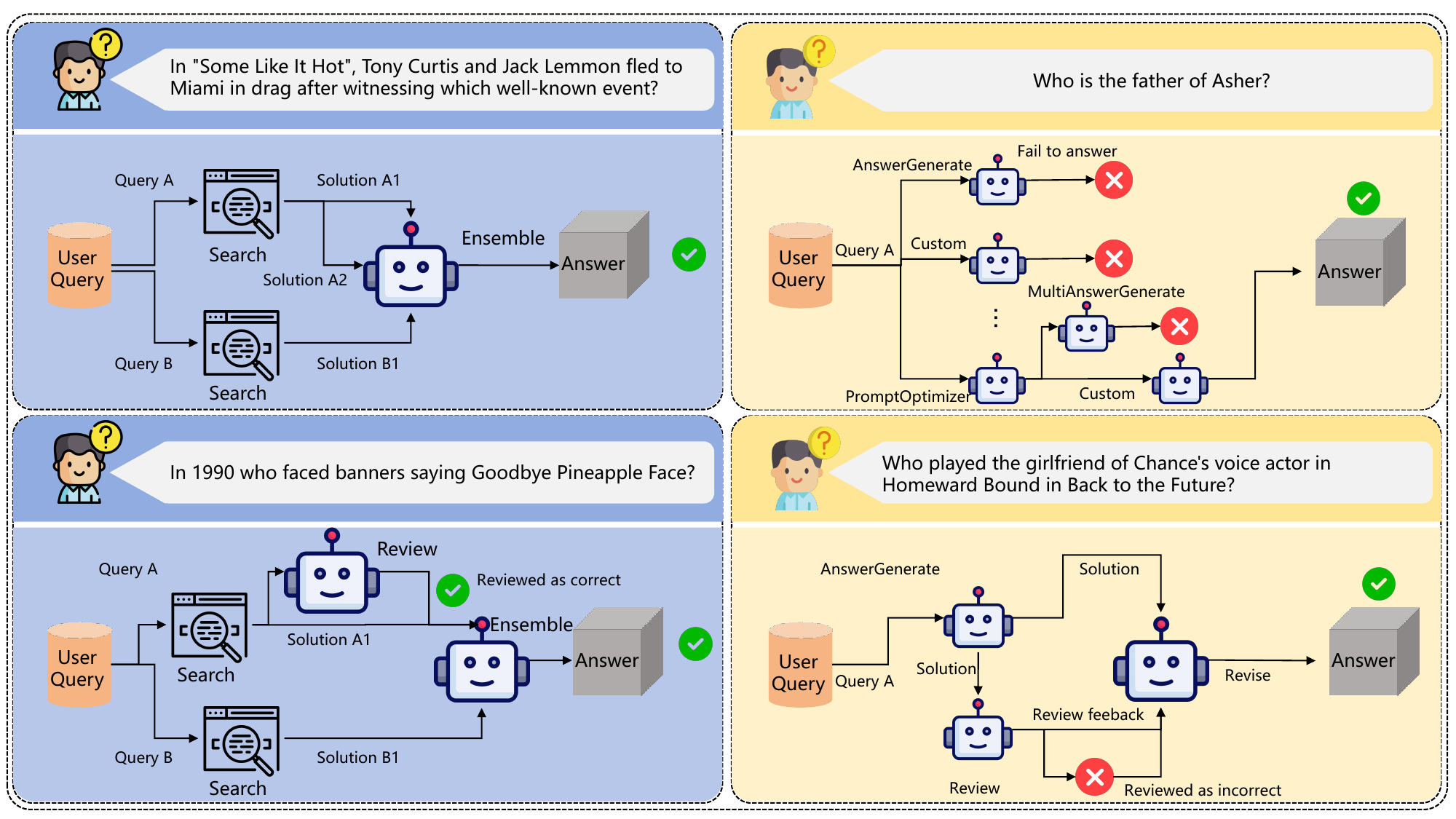}
    \caption{Visualization of workflow construction for \algs~(left, blue) and \alg~(right, yellow).}
    \label{fig:case_study}

\vspace{-5mm}
\end{figure*}

We establish baselines on Qwen2.5-32B-Instruct \cite{2025Qwen25} using \textbf{Direct Inference}, \textbf{Chain-of-Thought (CoT)} \cite{wei2022chain}, \textbf{Self-Consistency (SC)} \cite{wang2022self} with 5 paths, and \textbf{MedPrompt} \cite{nori2023can}. Regarding workflow optimization, we compare \alg~against \textbf{AFlow} \cite{aflow2024} and \textbf{MaAS} \cite{zhang2025multi}, which represent state-of-the-art task-level and query-level methods respectively. The detailed configurations for these baselines are provided in Appendix \ref{Appx:aflow_maas}. Finally, we benchmark \algs~against both PPO \cite{schulman2017proximal} and GRPO \cite{shao2024deepseekmath} variants of \textbf{Search-R1} \cite{jin2025search} to evaluate the effectiveness of our workflow optimization framework in search-augmented scenarios.

\vspace{-2mm}
\subsection{Experiments Details}
\label{sec:implementation}

\paragraph{Model Configuration.}
We employ Qwen2.5-7B-Instruct \cite{2025Qwen25} as the backbone for our workflow construction agent. We utilize a fixed Qwen2.5-32B-Instruct \cite{2025Qwen25} as the execution model across all agentic workflow methods, including AFlow and MaAS. Furthermore, we evaluate the performance using DeepSeek V3.2 \cite{liu2025deepseek} as the execution model in the ablation studies presented in Section \ref{sec:ablation}. 

\textbf{Training Setup.} We perform full-parameter fine-tuning to train two policy variants, \alg~and \algs~, which differ strictly in their available operator pools. Unless otherwise specified, we employ GSsPO as the default RL algorithm. Table \ref{tab:baseline_results} reports the comprehensive benchmarks validating our framework in both standard and search-augmented scenarios, while Table \ref{tab:workflow_performance} presents ablation studies comparing GSsPO against GRPO and GSPO. For detailed hyperparameters and configuration specifics, please refer to Appendix~\ref{Appx:training_details}.
\vspace{-1mm}

\subsection{Experiments Analysis}

\textbf{Superiority in Agentic Workflow Generation.} 
\alg~achieves state-of-the-art performance in the standard setting, significantly outperforming both inference-time prompting baselines and static workflow optimization methods such as AFlow and MaAS. This confirms that our dynamic optimization paradigm is fundamentally more effective for reasoning tasks than static approaches. Furthermore, when equipped with the Search operator, the performance is elevated to a new level: \algs~achieves the highest scores across all baselines, establishing the SOTA on the evaluated benchmarks with an average EM of 0.507.

\textbf{Case Study.}
Figure~\ref{fig:case_study} visualizes constructed workflows that reveal the emergence of \textbf{deep reflective capabilities}, where the agent transcends rigid execution sequences by continuously monitoring operator execution results to dynamically refine its strategies. This is exemplified when the agent refrains from accepting uncertain search results at face value, instead autonomously invoking the Review operator to validate them. Moreover, upon encountering execution failures, the agent immediately adjusts its plan to attempt alternative strategies, such as leveraging PromptOptimizer to reformulate the query expression and subsequently invoking other operators to re-attempt the solution. This rational persistence enables the agent to self-correct and construct optimal solution paths for complex queries. For further demonstrations of the raw responses, please refer to Appendix~\ref{subsec:more_case_studies}.
\vspace{-1mm}
\section{Ablation Study}
\label{sec:ablation}

\textbf{Robustness with Advanced Backbone.} We evaluated robustness by upgrading the backbone to DeepSeek V3.2 for both direct QA and workflow execution. This substitution yielded substantial performance gains across all agentic workflow methods. Notably, the performance advantage of automated workflow optimization over basic prompting baselines such as SC and MedPrompt increased significantly. These results indicate that superior execution models are critical for fully leveraging complex workflows to achieve qualitative performance breakthroughs.

\textbf{Different Reinforcement Learning Algorithms.} As shown in Table~\ref{tab:workflow_performance}, GSsPO consistently outperforms GRPO across all datasets and surpasses GSPO on the majority of tasks, validating the sub-sequence as the optimal optimization unit. Furthermore, Figure~\ref{fig:convergence_comparison} confirms that GSsPO maintains stable training dynamics and achieves efficient convergence.
\vspace{-1mm}
\section{Related Work}

\textbf{Automated Workflow Optimization.} Agentic evolution spans three paradigms. Modular frameworks like DSPy \cite{khattab2023dspy} and TextGrad \cite{yuksekgonul2024textgrad} use textual feedback, while GPTSwarm \cite{zhuge2024gptswarm} optimizes connectivity. Code-centric ADAS \cite{hu2024automated} programs architectures via meta-agents, extended by AFlow \cite{aflow2024} using MCTS. Query-adaptive methods like MAS-GPT \cite{ruimasgpt}, MaAS \cite{zhang2025multi}, and FlowReasoner \cite{gao2025flowreasoner} dynamically generate structures per query. However, most remain limited by a "Static Execution Trap," where pre-fixed structures cannot adapt to intermediate observations.

\vspace{-1mm}

\textbf{Reinforcement Learning for LLMs.} RL aligns LLMs via methods ranging from RLHF \cite{ouyang2022training} (e.g., PPO \cite{schulman2017proximal}) to direct optimization like DPO \cite{rafailov2023direct} and RLOO \cite{ahmadian2024back}, which often face off-policy reasoning challenges \cite{pang2024iterative}. Recently, "Pure RL" paradigms like DeepSeek-R1 \cite{guo2025deepseek} and GRPO \cite{shao2024deepseekmath} eliminated critics via group rewards, with stability enhanced by DAPO \cite{yu2025dapo}. However, a granularity mismatch persists in multi-turn scenarios: GRPO's token-level updates risk disrupting semantic integrity, while GSPO's \cite{zheng2025group} sequence-level signal remains too coarse for precise credit assignment.
\vspace{-1mm}
\section{Conclusion}
In this paper, we present \alg~, a framework that fundamentally transforms workflow construction from static synthesis into a dynamic sequential decision process. To enable this query-adaptive capability, we introduce Group Sub-sequence Policy Optimization (GSsPO). By establishing the Think-Action pair as the decision unit, GSsPO effectively resolves the optimization granularity mismatch inherent in conventional approaches. Extensive evaluations demonstrate that \alg~ significantly outperforms advanced baselines across seven benchmarks. Crucially, our analysis reveals the emergence of deep reflective capabilities, allowing the agent to autonomously self-correct during execution. Ultimately, Workflow-R1 establishes a robust foundation for autonomous agents, showcasing the vast potential of dynamic workflow optimization in the field of complex reasoning.

\section*{Impact Statement}
This paper presents work whose goal is to advance the field of Machine Learning. There are many potential societal consequences of our work, none which we feel must be specifically highlighted here.

\bibliography{example_paper}
\bibliographystyle{icml2026}

\newpage
\appendix
\onecolumn

\section{Experimental Setups}
\label{Appx:experimental_setups}

\subsection{Datasets}

We construct a mixed training set of 10,000 samples from NQ (4,618) and HotpotQA (5,382) to balance single-hop and multi-hop reasoning capabilities. For evaluation, we use seven datasets across two domains: General QA (NQ, TriviaQA, PopQA) and Multi-Hop QA (HotpotQA, 2WikiMultiHopQA, Musique, Bamboogle). The detailed statistics are shown in Table~\ref{tab:dataset_stats}.

\begin{table}[h]
    \centering
    \begin{tabular}{llrr}
        \toprule
        \textbf{Category} & \textbf{Dataset} & \textbf{Train} & \textbf{Test} \\
        \midrule
        \multirow{3}{*}{General QA} 
            & NQ & 4,618 & 3,610 \\
            & TriviaQA & -- & 11,313 \\
            & PopQA & -- & 14,267 \\
        \midrule
        \multirow{4}{*}{Multi-Hop QA} 
            & HotpotQA & 5,382 & 7,405 \\
            & 2WikiMultiHopQA & -- & 12,576 \\
            & Musique & -- & 2,417 \\
            & Bamboogle & -- & 125 \\
        \midrule
        \multicolumn{2}{l}{\textbf{Total}} & \textbf{10,000} & \textbf{51,713} \\
        \bottomrule
    \end{tabular}
    \caption{Dataset Statistics.}
    \label{tab:dataset_stats}
\end{table}

\subsection{Workflow Baselines}
\label{Appx:aflow_maas}

For the agentic workflow baselines, MaAS and AFlow represent state-of-the-art query-level and task-level workflow optimization methods, respectively. In our reproduction, we follow their official implementations: MaAS employs GPT-4o-mini \cite{hurst2024gpt} and AFlow employs Claude-3.5-Sonnet \cite{sonnet2024anthropic} as their respective workflow optimization models. To ensure fair comparison, we use Qwen2.5-32B-Instruct as the unified workflow execution model for both training and testing in main experiments.

\subsection{Training Configuration}
\label{Appx:training_details}

We utilize \textbf{Qwen2.5-7B-Instruct} as the policy model (workflow optimization model) and \textbf{Qwen2.5-32B-Instruct} as the workflow execution model for \alg~. All experiments perform full-parameter fine-tuning with the GSsPO algorithm using the VeRL framework~\cite{sheng2025hybridflow} on 2 NVIDIA H20 GPUs. The detailed hyperparameters are listed in Table~\ref{tab:hyperparameters}.

\begin{table}[h]
    \centering
    \begin{tabular}{lc}
        \toprule
        \textbf{Hyperparameter} & \textbf{Value} \\
        \midrule
        Training Epochs & 2 \\
        Training Samples & 10,000 \\
        Global Batch Size & 512 \\
        Rollout & 8 \\
        Total Training Steps & 38 \\
        Learning Rate & 1e-6 \\
        KL Coefficient & 0.001 \\
        Clip ratio & 0.2 \\
        \bottomrule
    \end{tabular}
    \caption{Training Hyperparameters for \alg~and \algs~.}
    \label{tab:hyperparameters}
\end{table}

Regarding the policy variants, \algs~replaces the \texttt{MultiAnswerGenerate} operator with the \texttt{Search} operator to enable external information retrieval, while \alg~retains the standard operator pool for self-contained reasoning.

\section{Policy Gradient of More RL}
\label{Appx:policy_gradient}

In this section, we provide the detailed gradient derivations for GRPO and GSPO to facilitate the comparative analysis.

\subsection{GRPO}
Group Relative Policy Optimization (GRPO) operates at the \textbf{token-level}. 

\textbf{Optimization Objective:}
The objective function is defined as follows (omitting the clipping mechanism for brevity):
\begin{equation}
    \mathcal{J}_{GRPO}(\theta) = \mathbb{E}_{x \sim \mathcal{D}, \{y_i\}_{i=1}^G \sim \pi_{\theta_{old}}} \left[ \frac{1}{G} \sum_{i=1}^G \frac{1}{|y_i|} \sum_{t=1}^{|y_i|} w_{i,t}(\theta) \hat{A}_{i,t} \right]
\end{equation}

\textbf{Importance Ratio Definition:}
The token-level importance ratio $w_{i,t}(\theta)$ is defined as:
\begin{equation}
    w_{i,t}(\theta) = \frac{\pi_\theta(y_{i,t}|x, y_{i,<t})}{\pi_{\theta_{old}}(y_{i,t}|x, y_{i,<t})}
\end{equation}

\textbf{Gradient Derivation:}
The policy gradient is derived by differentiating the objective function:
\begin{equation}
\begin{aligned}
    \nabla_\theta \mathcal{J}_{GRPO}(\theta) &= \nabla_\theta \mathbb{E} \left[ \frac{1}{G} \sum_{i=1}^G \frac{1}{|y_i|} \sum_{t=1}^{|y_i|} \frac{\pi_\theta(y_{i,t}|x, y_{i,<t})}{\pi_{\theta_{old}}(y_{i,t}|x, y_{i,<t})} \hat{A}_{i,t} \right] \\
    &= \mathbb{E} \left[ \frac{1}{G} \sum_{i=1}^G \frac{1}{|y_i|} \sum_{t=1}^{|y_i|} \hat{A}_{i,t} \nabla_\theta \left( \frac{\pi_\theta(y_{i,t}|x, y_{i,<t})}{\pi_{\theta_{old}}(y_{i,t}|x, y_{i,<t})} \right) \right] \\
    &= \mathbb{E} \left[ \frac{1}{G} \sum_{i=1}^G \frac{1}{|y_i|} \sum_{t=1}^{|y_i|} \frac{\pi_\theta(y_{i,t}|x, y_{i,<t})}{\pi_{\theta_{old}}(y_{i,t}|x, y_{i,<t})} \hat{A}_{i,t} \nabla_\theta \log \pi_\theta(y_{i,t} | x, y_{i,<t}) \right]
\end{aligned}
\end{equation}
This derivation confirms that GRPO updates the policy based on individual token likelihoods, weighted by their specific token-level importance ratios.

\subsection{GSPO}
Group Sequence Policy Optimization (GSPO) treats the \textbf{entire response} as a single unit.

\textbf{Optimization Objective:}
The objective function is formulated as follows (omitting the clipping mechanism for brevity):
\begin{equation}
    \mathcal{J}_{GSPO}(\theta) = \mathbb{E}_{x \sim \mathcal{D}, \{y_i\}_{i=1}^G \sim \pi_{\theta_{old}}} \left[ \frac{1}{G} \sum_{i=1}^G s_i(\theta) \hat{A}_i \right]
\end{equation}

\textbf{Importance Ratio Definition:}
The sequence-level importance sampling ratio $s_i(\theta)$ is defined based on the geometric mean of likelihood ratios:
\begin{equation}
    s_i(\theta) = \left( \frac{\pi_\theta(y_i|x)}{\pi_{\theta_{old}}(y_i|x)} \right)^{\frac{1}{|y_i|}}
\end{equation}

\textbf{Gradient Derivation:}
The gradient derivation proceeds as follows:
\begin{equation}
\begin{aligned}
    \nabla_\theta \mathcal{J}_{GSPO}(\theta) &= \nabla_\theta \mathbb{E} \left[ \frac{1}{G} \sum_{i=1}^G \left( \frac{\pi_\theta(y_i|x)}{\pi_{\theta_{old}}(y_i|x)} \right)^{\frac{1}{|y_i|}} \hat{A}_i \right] \\
    &= \mathbb{E} \left[ \frac{1}{G} \sum_{i=1}^G \hat{A}_i \nabla_\theta \left( \left( \frac{\pi_\theta(y_i|x)}{\pi_{\theta_{old}}(y_i|x)} \right)^{\frac{1}{|y_i|}} \right) \right] \\
    &= \mathbb{E} \left[ \frac{1}{G} \sum_{i=1}^G \left( \frac{\pi_\theta(y_i|x)}{\pi_{\theta_{old}}(y_i|x)} \right)^{\frac{1}{|y_i|}} \hat{A}_i \nabla_\theta \log \left( \left( \frac{\pi_\theta(y_i|x)}{\pi_{\theta_{old}}(y_i|x)} \right)^{\frac{1}{|y_i|}} \right) \right] \\
    &= \mathbb{E} \left[ \frac{1}{G} \sum_{i=1}^G  \frac{1}{|y_i|} \sum_{t=1}^{|y_i|} \left( \frac{\pi_\theta(y_i|x)}{\pi_{\theta_{old}}(y_i|x)} \right)^{\frac{1}{|y_i|}} \hat{A}_i \nabla_\theta \log \pi_\theta(y_{i,t} | x, y_{i,<t}) \right]
\end{aligned}
\end{equation}
In contrast to GRPO, GSPO applies a unified sequence-level weight to all tokens within a response, ensuring that the gradient updates are consistent with the semantic integrity of the sequence.

\section{More Results}
\label{Appx:more_details}

In addition to the main results using the 32B model, we also evaluate the performance of a smaller backbone, \textbf{Qwen2.5-7B-Instruct}, across the benchmarks. Table \ref{tab:baseline_results_7B} presents the performance on  four baseline methods.

\begin{table*}[h]
\centering
\resizebox{\textwidth}{!}{%
\begin{tabular}{l c ccc cccc c}
\toprule
\multirow{2}{*}{\textbf{Methods}} & \multirow{2}{*}{\textbf{Multi-turn}} & \multicolumn{3}{c}{\textbf{General QA}} & \multicolumn{4}{c}{\textbf{Multi-Hop QA}} & \multirow{2}{*}{\textbf{Avg.}} \\
\cmidrule(lr){3-5} \cmidrule(lr){6-9}
 & & NQ$^\dagger$ & TriviaQA & PopQA & HpQA$^\dagger$ & 2wiki & Musique & Bamb & \\
\midrule

\multicolumn{10}{l}{\textbf{Qwen2.5-7B-Instruct}} \\
\quad Direct      & \xmark & 0.030 & 0.051 & 0.056 & 0.098 & 0.184 & 0.011 & 0.048 & 0.068 \\
\quad CoT         & \xmark & 0.122 & 0.102 & 0.121 & 0.173 & 0.202 & 0.043 & 0.312 & 0.154 \\
\quad SC (CoT$\times$5) & \xmark & 0.142 & 0.110 & 0.130 & 0.188 & 0.216 & 0.050 & 0.336 & 0.167 \\
\quad MedPrompt   & \xmark & 0.114 & 0.104 & 0.119 & 0.146 & 0.155 & 0.046 & 0.320 & 0.143 \\

\bottomrule
\end{tabular}%
}
\caption{Performance comparison of baseline methods on benchmarks. The \textbf{Multi-turn} column indicates whether the method involves iterative interaction. $^\dagger$ represents in-domain datasets.}
\label{tab:baseline_results_7B}
\vspace{-6mm}
\end{table*}

\section{Details of Operator Description and Prompt Template}
\label{Appx:prompt_template}

\subsection{prompt template}

We present the training prompt template, which specifies multi-turn protocols and strict formatting constraints, alongside the complete operator pool and invocation syntax. Notably, we introduce the {\color{olive}WarningOp} as a safeguard against formatting violations (e.g., incorrect Think-Action structure). This operator intercepts invalid outputs and returns a warning message as observation feedback, explicitly guiding the agent to self-correct in subsequent steps.

\begin{mycolorbox}{Query}{\textbf{Query:}Prompt Template}
Answer the given question. Every time you get new information from an operator, you must conduct reasoning inside {\color{orange}$<$think$>$} and {\color{orange}$<$/think$>$} before taking any further action. This means you should have multiple thinking sessions - one after each time you receive information from operators, not just one initial thinking session. \\

After reasoning, if you find you can utilize an operator to help you increase the likelihood of obtaining a correct answer, you can call an operator according to the operator description below. And when you call the operator, you MUST follow the corresponding format. \\

Available Operators: \{AVAILABLE\_OPERATOR\_POOL\} \\

Before each operator call, you must explicitly reason inside {\color{orange}$<$think$>$} and {\color{orange}$<$/think$>$} about why you're choosing this operator, how you'll use it, and what you expect to learn.
When you call an operator, the response will be returned between {\color{teal}$<$info$>$} and {\color{teal}$<$/info$>$}. After receiving this response, you must again think inside {\color{orange}$<$think$>$} and {\color{orange}$<$/think$>$} about what the information means, how it helps answer the original question, and what your next step should be. \\

CRITICAL INSTRUCTION: Do not settle for simple, single-step operator calls. You are REQUIRED to aggressively explore diverse, complex, and unconventional combinations of operators. Your goal is not just to answer, but to exhaustively explore how different operators can synergize. You must push the boundaries of what these operators can achieve together. Passive, lazy, or simplistic strategies are unacceptable; you must demonstrate deep reasoning through rich, multi-step operator orchestration to uncover their unique capabilities and "magical" combined effects. \\

IMPORTANT CONSTRAINT: You have a MAXIMUM of 20 operator calls. Plan your strategy wisely - you must gather sufficient information and reach a confident answer within this limit. \\

Always follow this cycle: think, then optionally use an operator if needed, then when you receive information, think again, then decide your next step. Repeat this cycle until you have enough information to answer confidently.
After you have gathered enough information and conducted deep reasoning through multiple thinking cycles, if you are truly certain you can answer the query correctly, directly provide your final answer in only a few words or a short phrase inside {\color{red}$<$answer$>$} and {\color{red}$<$/answer$>$} to exactly answer the query. If you're not certain, continue gathering information and reasoning until you reach certainty. \\

Query: \{USER\_QUERY\}
    
\end{mycolorbox}

\vspace{-1mm}

\begin{mycolorbox}{Query}{\textbf{Operators Description and Usage}}
    
\textbf{\color{blue}AnswerGenerate}: \\
\textbf{Description}: Call an advanced LLM to think and answer questions. Useful for domain-specific questions, knowledge QA, reasoning tasks, etc. \\
\textbf{Usage}: {\color{blue}$<$tool$>$}AnswerGenerate: YOUR\_QUESTION{\color{blue}$<$/tool$>$}

\vspace{0.2cm}

\textbf{\color{blue}Custom}: \\
\textbf{Description}: Call an LLM to execute any task you define. As long as you provide a clear and detailed instruction along with the task content, the LLM can accomplish virtually anything you need. \\
\textbf{Usage}: {\color{blue}$<$tool$>$}Custom: $<$task$>$YOUR\_TASK\_CONTENT$<$/task$>$ $<$instruction$>$YOUR\_INSTRUCTION
$<$/instruction$>$ {\color{blue}$<$/tool$>$}

\vspace{0.2cm}

\textbf{\color{blue}PromptOptimizer}: \\
\textbf{Description}: Call an LLM to refine and optimize a query's clarification by adding details and creating a step-by-step plan. Useful when the original query is ambiguous or needs more context. \\
\textbf{Usage}: {\color{blue}$<$tool$>$}PromptOptimizer: YOUR\_QUERY{\color{blue}$<$/tool$>$}

\vspace{0.2cm}

\textbf{\color{blue}Review}: \\
\textbf{Description}: Call an LLM to critically review a solution's correctness and provide feedback. Useful when you want to verify whether some answer is correct before finalizing. \\
\textbf{Usage}: {\color{blue}$<$tool$>$}Review: $<$question$>$ORIGINAL\_QUESTION$<$/question$>$$<$solution$>$SOLUTION\_TO\_REVIEW
$<$/solution$>$ {\color{blue}$<$/tool$>$}

\vspace{0.2cm}

\textbf{\color{blue}Revise}: \\
\textbf{Description}: Call an LLM to revise a solution based on feedback. Useful when you have received critical feedback indicating there are something wrong in the solution. \\
\textbf{Usage}: {\color{blue}$<$tool$>$}Revise: $<$question$>$ORIGINAL\_QUESTION$<$/question$>$$<$solution$>$SOLUTION\_TO\_REVISE
$<$/solution$>$ $<$feedback$>$FEEDBACK\_RECEIVED$<$/feedback$>$ {\color{blue}$<$/tool$>$}

\vspace{0.2cm}

\textbf{\color{blue}MultiAnswerGenerate}: \\
\textbf{Description}: Call an advanced LLM to generate multiple independent answers (3 times) for the same question. Useful when you want to explore diverse reasoning paths or prepare candidates for ensemble selection. \\
\textbf{Usage}: {\color{blue}$<$tool$>$}MultiAnswerGenerate: YOUR\_QUESTION{\color{blue}$<$/tool$>$}

\vspace{0.2cm}

\textbf{\color{blue}ScEnsemble}: \\
\textbf{Description}: Call an LLM to select the best answer from multiple candidate solutions through self-consistency ensemble selection. You must provide at least two candidate solutions. \\
\textbf{Usage}: {\color{blue}$<$tool$>$}ScEnsemble: $<$query$>$ORIGINAL\_QUESTION$<$/query$>$$<$solution$>$CANDIDATE\_1$<$/solution$>$
...$<$solution$>$CANDIDATE\_N$<$/solution$>$ {\color{blue}$<$/tool$>$}

\vspace{0.2cm}

\textbf{\color{blue}Search}: \\
\textbf{Description}: Call a search engine to search for information on the web. Useful when you need to look up facts that are not already available to you. \\
\textbf{Usage}: {\color{blue}$<$tool$>$}Search: YOUR\_QUERY{\color{blue}$<$/tool$>$}

\vspace{0.2cm}


\textbf{\color{olive}WarningOp}: \\
\textbf{Description}: A special operator that returns a warning message when the model fails to follow the required output format. \\
\textbf{Output}: WARNING!!! I did not follow the required format. If I want to use an operator, I should use $<$tool$>$OperatorName: input$<$/tool$>$. If I want to give the final answer, I should use $<$answer$>$...$<$/answer$>$.

\end{mycolorbox}

\section{More Case Studies}
\label{subsec:more_case_studies}

In this section, we present a comprehensive set of qualitative examples, comprising four cases utilizing \algs~and four cases utilizing \alg~. These case studies underscore several compelling emergent behaviors inherent to our framework. We observe that when an operator fails to return valid information due to internal errors, the agent avoids stagnation, instead intelligently pivoting to alternative strategies to resolve the query. Furthermore, we identify robust self-correction mechanisms; upon triggering format warnings, the agent explicitly reflects on the error within its reasoning traces (inside \texttt{<think>} tags). Collectively, these emergent behaviors attest to the practical value of our proposed paradigm, marking a bold exploration into the frontier of autonomous multi-turn workflow construction.

\begin{mycolorbox}{Query}{\textbf{Query:} \texttt{What would Kevin Scale have been called if it had adopted the originator's first name?} \\ \\
\textbf{Ground Truth:} \texttt{William Scale}}

[ASSISTANT] \\
{\color{orange}$<$think$>$}First, I need to understand who Kevin Scale is and what the originator's first name would be in this context. However, the query does not provide enough context about Kevin Scale or the originator. I will start by searching for information about Kevin Scale to understand its context better.{\color{orange}$<$/think$>$} \\
{\color{blue}$<$tool$>$}Search: What is Kevin Scale and what does it refer to?{\color{blue}$<$/tool$>$} \\
\noindent\rule[0.5ex]{\textwidth}{0.5pt}

[OPERATOR] \\
{\color{teal}$<$info$>$}Search results for 'What is Kevin Scale and what does it refer to?': \\ 
Title: Kelvin \\
Snippet: The Kelvin scale is an absolute temperature scale that starts at the lowest possible temperature (absolute zero), taken to be 0 K. By definition, the Celsius ... \\ \\
Title: Kelvin Temperature Scale Definition \\
Snippet: The Kelvin temperature scale is used by scientists because they wanted a temperature scale where zero reflects the complete absence of thermal energy. \\ \\
Title: Kelvin (K) | Definition \& Facts \\
Snippet: The ... ... by shifting the Celsius scale by $-273.15^\circ$ so that absolute zero coincides with 0 K) ... \\ \\
Title: Kelvin Scale - an overview | ScienceDirect Topics \\
Snippet: The Kelvin scale is defined as a temperature scale that extends the degree Celsius scale down to absolute zero (0 K), representing a complete absence of ... \\ \\
Title: What is the Kelvin Temperature Scale? \\
Snippet: Outlining what the Kelvin temperature scale is and how it is used in chemistry. Kelvin and degrees celsius are compared and absolute zero is ...{\color{teal}$<$/info$>$}\\
\noindent\rule[0.5ex]{\textwidth}{0.5pt}

[ASSISTANT] \\
{\color{orange}$<$think$>$}The search results do not provide any direct information about Kevin Scale. However, the results are mostly about the Kelvin temperature scale, which might be a misinterpretation of the query. I need to clarify the context of 'Kevin Scale' and the originator's first name. I will use the PromptOptimizer to refine the query for better understanding.{\color{orange}$<$/think$>$} \\
{\color{blue}$<$tool$>$}PromptOptimizer: What would Kevin Scale have been called if it had adopted the originator's first name?{\color{blue}$<$/tool$>$} \\
\noindent\rule[0.5ex]{\textwidth}{0.5pt}

[OPERATOR] \\
{\color{teal}$<$info$>$}PromptOptimizer thought process: The query is asking about a hypothetical scenario where a concept or theory named \"Kevin Scale\" has its name changed based on the originator's first name. However, there seems to be a misunderstanding in the query as \"Kevin Scale\" does not appear to be a known scientific or theoretical concept. It might be a fictional or misnamed concept. To refine this query, we need to clarify what \"Kevin Scale\" refers to and who the originator is. If we assume that \"Kevin Scale\" is a plac... ...'s first name instead. For example, if the concept was originally named after a person with the full name \"John Kevin,\" and it was currently called \"Kevin Scale,\" the query wants to know what it would be called if it were named after John instead. \\
PromptOptimizer answer: Assuming \"Kevin Scale\" is a placeholder for a real concept named after an individual, the refined query would be: If the concept currently known as \"Kevin Scale\" was instead named after the originator's first name, what would it be called?{\color{teal}$<$/info$>$} \\
\noindent\rule[0.5ex]{\textwidth}{0.5pt}

[ASSISTANT] \\
{\color{orange}$<$think$>$}The PromptOptimizer provided a more specific and clear version of the query. However, it still does not give a clear answer about what \"Kevin Scale\" refers to. I need to search for a more specific context of \"Kevin Scale\" to understand its origin and the originator's first name. I will now search for \"Kevin Scale\" to find out what it is and who the originator is.{\color{orange}$<$/think$>$} \\
{\color{blue}$<$tool$>$}Search: What is Kevin Scale and who is the originator?{\color{blue}$<$/tool$>$} \\
\noindent\rule[0.5ex]{\textwidth}{0.5pt}

[OPERATOR] \\
{\color{teal}$<$info$>$}Search results for 'What is Kevin Scale and who is the originator?': \\
Title: Kelvin \\
Snippet: The 19th-century British scientist Lord Kelvin first developed and proposed the scale. It was often called the \"absolute Celsius\" scale in the early 20th ... \\ \\
Title: Celsius and Kelvin scales | Research Starters \\
Snippet: The Kelvin scale is another common temperature scale. British inventor and scientist William Thomson, also known as Lord Kelvin, developed the scale in 1848. \\ \\
Title: Thomson, William$<$br$>$(Lord Kelvin... ... but is best remembered as the originator of the kelvin* temperature scale. In 1848 Thomson proposed a ... \\ \\
Title: Lord Kelvin \\
Snippet: ... inventor Wilhelm Röntgen. ... Instead, he was named Lord Kelvin, a name that now lives on as the Kelvin scale for absolute temperature, an ... \\ \\
Title: How was Kelvin discovered and named after Lord Kelvin? \\
Snippet: In 1848, British mathematician and scientist William Thomson (also known as Lord Kelvin) proposed an absolute temperature scale where the lowest possible ...{\color{teal}$<$/info$>$} \\
\noindent\rule[0.5ex]{\textwidth}{0.5pt}

[ASSISTANT] \\
{\color{orange}$<$think$>$}The search results indicate that \"Kelvin\" is the scale named after Lord Kelvin, a 19th-century British scientist. The query seems to be asking about a hypothetical scenario where the scale is named after the originator's first name instead of the full name \"Kelvin.\" Since the originator is Lord Kelvin, and his first name is William, the scale would have been called the \"William Scale\" if it had adopted the originator's first name. Therefore, I can now answer the query.{\color{orange}$<$/think$>$} \\
{\color{red}$<$answer$>$}William Scale{\color{red}$<$/answer$>$}
\end{mycolorbox}


\begin{mycolorbox}{Query}{
    \textbf{Query:} \texttt{In 1990 who faced banners saying Goodbye Pineapple Face?} \\ \\
    \textbf{Ground Truth:} \texttt{["Manuel Antonio Noriega Moreno", "Manuel Noreiga", ...]}
}

[ASSISTANT] \\
{\color{orange}$<$think$>$}First, I need to find out who faced banners saying "Goodbye Pineapple Face" in 1990. I will start with a search query.{\color{orange}$<$/think$>$} \\
{\color{blue}$<$tool$>$}Search: In 1990 who faced banners saying Goodbye Pineapple Face{\color{blue}$<$/tool$>$} \\
\noindent\rule[0.5ex]{\textwidth}{0.5pt}

[OPERATOR] \\
{\color{teal}$<$info$>$}Search results for 'In 1990 who faced banners saying Goodbye Pineapple Face': \\
Title: World Leaders Trivia Quiz \\
Snippet: In 1990 who faced banners saying Goodbye Pineapple Face? remove\_red\_eye Show Answer. 20. Who was the youngest US Vice President of the 20th century ... \\ \\
Title: Manuel Noriega \\
Snippet: Manuel Antonio Noriega Moreno was a Panamanian military officer and politician who was the de facto ruler of Panama from 1983 to 1989. \\ \\
Title: Harvesting Prosperity: Technology and Productivity Growth \\
Snippet: ...   ... \\ \\
Title: Sustainable diets and biodiversity \\
Snippet: 1990 and 2015. In 2011, some countries remain far from reaching this target, and ensuring global food security persists as one of the greatest challenges of ... \\ \\
Title: The Adaptability of Marketing Systems to Interventions in ... \\
Snippet: by MH Hounhouigan · 2014 · Cited by 38 — When pineapple production intensified in the late 1990s in an effort to stimulate economic growth, the marketing sys- tem failed to sell the increasing ...{\color{teal}$<$/info$>$} \\
\noindent\rule[0.5ex]{\textwidth}{0.5pt}

[ASSISTANT] \\
{\color{orange}$<$think$>$}The search results are not very clear and direct. However, one snippet from "World Leaders Trivia Quiz" seems relevant as it directly asks about banners saying "Goodbye Pineapple Face" in 1990. I need to click on the link to get the answer. But since I can't click, I will assume the answer is provided in the quiz and it's likely a trivia question about a political figure or event. I will use the Review operator to verify if the context of the trivia question can be correctly interpreted.{\color{orange}$<$/think$>$} \\
{\color{blue}$<$tool$>$}Review: $<$question$>$In 1990 who faced banners saying Goodbye Pineapple Face?$<$question$>$ $<$solution$>$Manuel Noriega$<$/solution$>${\color{blue}$<$/tool$>$} \\
\noindent\rule[0.5ex]{\textwidth}{0.5pt}

[OPERATOR] \\
{\color{teal}$<$info$>$}Review feedback: The provided solution, "Manuel Noriega," does not seem to be correct based on the context of the question. The term "Pineapple Face" is not commonly associated with Manuel Noriega in historical records or media coverage related to him. \\
A more likely scenario involves former British Prime Minister Margaret Thatcher, who was sometimes referred to as "Dagwood Bumface" or faced similar derogatory banners during her tenure, but not specifically "Goodbye Pineapple Face." It seems there might be a mix-up or misinterpretation of the event or person involved. \\
To improve the solution, it would be beneficial to research the specific event where someone faced banners with this exact phrase. This could involve looking into political figures from the early 1990s who were subjects of public protest or ridicule.{\color{teal}$<$/info$>$} \\
\noindent\rule[0.5ex]{\textwidth}{0.5pt}

[ASSISTANT] \\
{\color{orange}$<$think$>$}The review feedback suggests that my initial guess of Manuel Noriega was incorrect. The term "Pineapple Face" is not commonly associated with him. The feedback hints that the event might be related to a political figure from the early 1990s who faced such banners. Given the context, Margaret Thatcher seems a plausible candidate as she was a prominent political figure during that time and faced various forms of public protest. I will now use the Search operator to look for more specific information about banners saying "Goodbye Pineapple Face" in the context of Margaret Thatcher or similar political figures from the early 1990s.{\color{orange}$<$/think$>$} \\
{\color{blue}$<$tool$>$}Search: Who faced banners saying Goodbye Pineapple Face in the early 1990s{\color{blue}$<$/tool$>$} \\
\noindent\rule[0.5ex]{\textwidth}{0.5pt}

[OPERATOR] \\
{\color{teal}$<$info$>$}Search results for 'Who faced banners saying Goodbye Pineapple Face in the early 1990s': \\
Title: Manuel Noriega \\
Snippet: Manuel Antonio Noriega Moreno was a Panamanian military officer and politician who was the de facto ruler of Panama from 1983 to 1989. \\ \\
Title: Manuel Noriega. They both got pothole pineapple face. \\
Snippet: Mohamed looks just like this man - Manuel Noriega. They both got pothole pineapple face. · 15y · Public · Judyth E. Beerup and Rachel Kovach. \\ \\
Title: Manuel Noriega, Dictator Ousted by ... ...tives protesting the use of his likeness and the “Pineapple Face” moniker in a “Call of Duty” video game.) Embracing his ... \\ \\
Title: General Noriega sues Call Of Duty for portraying him as a ' ... \\
Snippet: Former Panama dictator General Noriega sues Call Of Duty makers for portraying him as a 'kidnapper, murderer and old pineapple face'. \\ \\
Title: Dirty Manuel \\
Snippet: The story of General Noriega's rise and fall inspired several good nonfiction books in the early 1990's, and it clearly delighted Lawrence ...{\color{teal}$<$/info$>$} \\
\noindent\rule[0.5ex]{\textwidth}{0.5pt}

[ASSISTANT] \\
{\color{orange}$<$think$>$}The search results consistently point towards Manuel Noriega and his association with the term "Pineapple Face." However, the specific banners saying "Goodbye Pineapple Face" seem to be a misunderstanding or misattribution. The feedback and search results suggest that the term "Pineapple Face" was used in a different context, possibly related to Margaret Thatcher or another political figure. Given the persistent association of "Pineapple Face" with Manuel Noriega and the lack of direct evidence for the specific banners in question, I need to re-evaluate the context and consider the possibility that the question might be referring to a different event or figure. Since the term "Pineapple Face" is strongly linked to Manuel Noriega, and the banners saying "Goodbye Pineapple Face" might be a misinterpretation, I will consider the context of the question and the feedback. The feedback suggests that the term "Pineapple Face" was used in a different context, possibly related to Margaret Thatcher. However, the direct association with banners saying "Goodbye Pineapple Face" is not clear. I will use the ScEnsemble operator to compare the two candidate solutions: Manuel Noriega and Margaret Thatcher, to see if the collective wisdom can help me decide.{\color{orange}$<$/think$>$} \\
{\color{blue}$<$tool$>$}ScEnsemble: $<$query$>$ In 1990 who faced banners saying Goodbye Pineapple Face?$<$query$>$ $<$solution$>$Manuel Noriega$<$/solution$>$$<$solution$>$Margaret Thatcher$<$/solution$>${\color{blue}$<$/tool$>$} \\
\noindent\rule[0.5ex]{\textwidth}{0.5pt}

[OPERATOR] \\
{\color{teal}$<$info$>$}ScEnsemble thought process: To analyze the provided options, we need to consider the historical context of the phrase "Goodbye Pineapple Face" and who it might refer to. \\
Option A: Manuel Noriega - Manuel Noriega was the de facto leader of Panama from 1983 until his arrest in 1989. He was later extradited to France and then to the United States to face charges. The term "Pineapple Face" could be a derogatory nickname given to him, possibly due to his appearance or as part of political satire or media portra... ... being the correct answer based on the information given. However, if we must choose between the two based on the likelihood of facing such banners due to their prominence in international affairs during that period, Manuel Noriega might be slightly more plausible due to the intense media scrutiny and political opposition he faced around the time of his extradition. \\
However, it's important to note that without concrete historical evidence, both options remain speculative. \\
ScEnsemble answer: Manuel Noriega{\color{teal}$<$/info$>$} \\
\noindent\rule[0.5ex]{\textwidth}{0.5pt}

[ASSISTANT] \\
{\color{orange}$<$think$>$}The ScEnsemble operator has provided a collective wisdom that Manuel Noriega is the more plausible answer based on the historical context and the intense media scrutiny and political opposition he faced around the time of his extradition. While the term "Pineapple Face" is strongly associated with Manuel Noriega, the specific banners saying "Goodbye Pineapple Face" might be a misinterpretation or a different event. However, given the context and the feedback, Manuel Noriega is the most likely answer. I will now provide the answer based on the collective wisdom.{\color{orange}$<$/think$>$} \\
{\color{red}$<$answer$>$}Manuel Noriega{\color{red}$<$/answer$>$}
\end{mycolorbox}

\begin{mycolorbox}{Query}{\textbf{Query:} \texttt{Which film has the director died earlier, Miracle Of Marcelino or The Accompanist?} \\ \\
\textbf{Ground Truth:} \texttt{["Miracle of Marcelino", "Miracle Of Marcelino"]}}

[ASSISTANT] \\
{\color{orange}$<$think$>$}First, I need to find out when the director of "Miracle Of Marcelino" died and when the director of "The Accompanist" died. I will start by searching for the director of "Miracle Of Marcelino". {\color{orange}$<$/think$>$} \\
{\color{blue}$<$tool$>$}Search: When did the director of Miracle Of Marcelino die?{\color{blue}$<$/tool$>$} \\
\noindent\rule[0.5ex]{\textwidth}{0.5pt}

[OPERATOR] \\
{\color{teal}$<$info$>$}Search results for 'When did the director of Miracle Of Marcelino die?': \\
Title: Miracle of Marcelino \\
Snippet: Miracle of Marcelino is a 1955 Spanish film written by José Maria Sanchez-Silva, based on his novel, and directed by Ladislao Vajda. \\ \\
Title: The Miracle of Marcelino (1955) \\
Snippet: The Miracle of Marcelino: Directed by Ladislao Vajda. With Rafael Rivelles, Antonio Vico, Juan Calvo, José Marco Davó. In 1950s Spain, a young orphan boy ... \\ \\
Title: 'The Mircale of Marcelino' or — Twelve Men and a Baby... ...miracle of Marcelino. Christ almighty indeed. Miracle Of Marcelino · Ladislao Vajda · Spanish Cinema. \\ \\
Title: Vajda, László (1906–1965) - NFI \\
Snippet: László Vajda, who captured the ominous atmosphere of the 1930s with novel candour, became director of Spain's most celebrated films after ... \\ \\
Title: Marcelino pan y vino, 1955) is a heartwarming Spanish film ... \\
Snippet: Marcelino falls asleep in Jesus' arms, dies peacefully ... 'Miracle of Marcelino' is a timeless Spanish classic that will touch your heart!{\color{teal}$<$/info$>$} \\
\noindent\rule[0.5ex]{\textwidth}{0.5pt}

[ASSISTANT] \\
{\color{orange}$<$think$>$}The search results indicate that Ladislao Vajda was the director of "Miracle Of Marcelino". Now I need to find out when he died. From the search results, I can see that Ladislao Vajda died in 1965. Next, I will search for the director of "The Accompanist" and when he died.{\color{orange}$<$/think$>$} \\
{\color{blue}$<$tool$>$}Search: When did the director of The Accompanist die?{\color{blue}$<$/tool$>$} \\
\noindent\rule[0.5ex]{\textwidth}{0.5pt}

[OPERATOR] \\
{\color{teal}$<$info$>$}Search results for 'When did the director of The Accompanist die?': \\
Title: (Updated) Vocal coach and accompanist Martin Isepp dies ... \\
Snippet: British accompanist, conductor and vocal coach Martin Isepp died on Christmas Day, aged 81. Here is the obituary notice, as published in The ... \\ \\
Title: James Levine, Former Met Opera Music Director, Is Dead ... \\
Snippet: Len Horovitz, confirmed his death to NPR, saying that Levine died of natural causes. He was 77 years old. Over four and a half decades, Levine ... ...composer, arranger, and educator who was best known and celebrated as an accompanist for jazz singers, died May 13 ... \\ \\
Title: James Levine \\
Snippet: Born, (1943-06-23)June 23, 1943. Cincinnati, Ohio, U.S. ; Died, March 9, 2021(2021-03-09) (aged 77). Palm Springs, California, U.S. ; Occupations, Conductor, ... \\ \\
Title: Martin Isepp, accompanist, has died \\
Snippet: Martin Isepp, who worked as an accompanist, conductor and vocal coach, died on Christmas Day. For several years he was associated with Glyndebourne.{\color{teal}$<$/info$>$} \\
\noindent\rule[0.5ex]{\textwidth}{0.5pt}

[ASSISTANT] \\
{\color{orange}$<$think$>$}The search results for "The Accompanist" do not directly provide the director's name or death date. However, the results for "Martin Isepp" and "James Levine" seem to be related to accompanists rather than directors. I need to clarify the director of "The Accompanist" and find out when he died. I will use the Custom operator to refine the search and get more specific results.{\color{orange}$<$/think$>$} \\
{\color{blue}$<$tool$>$}Custom: $<$task$>$Find the director of the film "The Accompanist" and when he died.$<$/task$>$$<$instruction$>$Search for the director of the film "The Accompanist" and the year of his death. Provide the director's name and the year of death.$<$/instruction$>${\color{blue}$<$/tool$>$} \\
\noindent\rule[0.5ex]{\textwidth}{0.5pt}

[OPERATOR] \\
{\color{teal}$<$info$>$}Custom answer: Director: Patrice Leconte; Year of Death: Still Alive (as of 2023){\color{teal}$<$/info$>$} \\
\noindent\rule[0.5ex]{\textwidth}{0.5pt}

[ASSISTANT] \\
{\color{orange}$<$think$>$}The Custom operator provided information that Patrice Leconte is the director of "The Accompanist" and he is still alive as of 2023. Therefore, the director of "The Accompanist" has not died yet. On the other hand, Ladislao Vajda, the director of "Miracle Of Marcelino", died in 1965. Thus, the film directed by the deceased director is "Miracle Of Marcelino".{\color{orange}$<$/think$>$} \\
{\color{red}$<$answer$>$}Miracle Of Marcelino{\color{red}$<$/answer$>$}
\end{mycolorbox}

\begin{mycolorbox}{Query}{\textbf{Query:} \texttt{In "Some Like It Hot", Tony Curtis and Jack Lemmon fled to Miami in drag after witnessing which well-known event?} \\ \\
\textbf{Ground Truth:} \texttt{["John May (gangster)","Valentines Day Massacre", ...]}
}

[ASSISTANT] \\
{\color{orange}$<$think$>$}First, I need to find out the well-known event that Tony Curtis and Jack Lemmon witnessed in the movie "Some Like It Hot" which made them flee to Miami in drag. I will start with a search query.{\color{orange}$<$/think$>$} \\
{\color{blue}$<$tool$>$}Search: In "Some Like It Hot", Tony Curtis and Jack Lemmon fled to Miami in drag after witnessing which well-known event?{\color{blue}$<$/tool$>$} \\
\noindent\rule[0.5ex]{\textwidth}{0.5pt}

[OPERATOR] \\
{\color{teal}$<$info$>$}Search results for 'In "Some Like It Hot", Tony Curtis and Jack Lemmon fled to Miami in drag after witnessing which well-known event?': \\
Title: Some Like It Hot \\
Snippet: Some Like It Hot is a 1959 American crime comedy film produced, co-written and directed by Billy Wilder. It stars Marilyn Monroe, Tony Curtis and Jack ... \\ \\
Title: Some Like It Hot (1959) \\
Snippet: Some Like It Hot ... After two male musicians witness a mob hit, they flee the state in an all-female band disguised as women, but further complic... ... two male musicians witness a mob hit, they flee the state in an all-female band ... \\ \\
Title: CONVERSATIONS WITH BILLY WILDER: Film's Fearless ... \\
Snippet: ... famous line delivered with cheerful gusto by Joe E. Brown at the conclusion of “Some Like it Hot” after Jack Lemmon's frantic character in drag ... \\ \\
Title: Proctors Theatre Archives \\
Snippet: On Thursday, September 19, my wife and I went to Proctors Theatre to see the national touring company production of Some Like It Hot. A couple of seasons ago, ...{\color{teal}$<$/info$>$} \\
\noindent\rule[0.5ex]{\textwidth}{0.5pt}

[ASSISTANT] \\
{\color{orange}$<$think$>$}The search results indicate that Tony Curtis and Jack Lemmon fled to Miami in drag after witnessing a mob hit in the movie "Some Like It Hot". I need to find out the specific name of this event. I will refine my search query to find the exact name of the event.{\color{orange}$<$/think$>$} \\
{\color{blue}$<$tool$>$}Search: What is the name of the mob hit witnessed by Tony Curtis and Jack Lemmon in the movie "Some Like It Hot"?{\color{blue}$<$/tool$>$} \\
\noindent\rule[0.5ex]{\textwidth}{0.5pt}

[OPERATOR] \\
{\color{teal}$<$info$>$}Search results for 'What is the name of the mob hit witnessed by Tony Curtis and Jack Lemmon in the movie "Some Like It Hot"?': \\
Title: Jack Lemmon and Tony Curtis in Some Like It Hot 1959 ... \\
Snippet: Jack Lemmon and Tony Curtis in Some Like It Hot 1959 After two male musicians witness a mob hit, they flee the state in an all-female band ... \\ \\
Title: Some Like it Hot (1959) \\
Snippet: Jack Lemmon and Tony Curtis dress in drag, and find themselves in some queer situations, in the classic 1959 comedy SOME LIKE ... ...ly Wilder, released in 1959, and is often regarded as one of the greatest comedies of all time. \\ \\
Title: SAT 07 JAN – 'Some Like It Hot' (12A) - West Side Cinema \\
Snippet: 'Some Like It Hot' (1959) Billy Wilder. When two male musicians witness a mob hit, they flee the state in an all-female band disguised as women, ... \\ \\
Title: Some Like It Hot \\
Snippet: Some Like It Hot. Two musicians witness a mob hit and struggle to find a way out of the city before they are found by the gangsters. Their only opportunity ...{\color{teal}$<$/info$>$} \\
\noindent\rule[0.5ex]{\textwidth}{0.5pt}

[ASSISTANT] \\
{\color{orange}$<$think$>$}The search results are still not providing a specific name for the mob hit. However, the context and the plot summary strongly suggest that the event they witnessed was a significant mob-related incident. Given the nature of the movie and the plot, it is likely that the event was a well-known mob-related event in the context of the film. Since the search results do not provide a specific name, and based on the context of the movie, the event they witnessed was a mob hit. The most plausible answer, given the context, is the St. Valentine's Day Massacre, a notorious event in the history of organized crime in Chicago. However, to be certain, I will use the ScEnsemble operator to compare multiple plausible answers.{\color{orange}$<$/think$>$} \\
{\color{blue}$<$tool$>$}ScEnsemble: $<$query$>$What is the name of the mob hit witnessed by Tony Curtis and Jack Lemmon in the movie "Some Like It Hot"?$<$query$>$ $<$solution$>$St. Valentine's Day Massacre$<$/solution$>$ $<$solution$>$Chicago Massacre$<$/solution$>$ $<$solution$>$Chicago Mob Hit$<$/solution$>${\color{blue}$<$/tool$>$} \\
\noindent\rule[0.5ex]{\textwidth}{0.5pt}

[OPERATOR] \\
{\color{teal}$<$info$>$}ScEnsemble thought process: To analyze the provided options, we need to consider the context of the movie "Some Like It Hot." The film is set in the 1920s during the Prohibition era, a time when organized crime was rampant, especially in cities like Chicago. The plot involves two musicians who witness a significant event involving gangsters and decide to flee disguised as women in an all-female band. \\
Option A suggests the "St. Valentine's Day Massacre," which is a real historical event that occurred on Feb... ...various events related to organized crime in Chicago. However, it does not provide enough detail to pinpoint a specific event that would be memorable enough to be referenced in a movie. \\
Considering the historical accuracy and the specificity of the event, Option A, "St. Valentine's Day Massacre," seems to be the most precise and relevant to the context of the movie. It is a well-known event that fits the timeframe and thematic elements of "Some Like It Hot." \\
ScEnsemble answer: St. Valentine's Day Massacre{\color{teal}$<$/info$>$} \\
\noindent\rule[0.5ex]{\textwidth}{0.5pt}

[ASSISTANT] \\
{\color{orange}$<$think$>$}The ScEnsemble operator has analyzed the options and concluded that the "St. Valentine's Day Massacre" is the most precise and relevant event that fits the context of the movie "Some Like It Hot." This aligns with the historical and thematic elements of the film, where a significant mob-related event would be a pivotal plot point. Therefore, I can confidently answer the query.{\color{orange}$<$/think$>$} \\
{\color{red}$<$answer$>$}St. Valentine's Day Massacre{\color{red}$<$/answer$>$}
\end{mycolorbox}

\begin{mycolorbox}{Query}{\textbf{Query:} \texttt{Who is the father of Asher?} \\ \\
\textbf{Ground Truth:} \texttt{["Jacob", "Israel", "Ya'akov", "Ya'aqov", "Yaakov", "Yisrael"]}}

[ASSISTANT] \\
{\color{orange}$<$think$>$}To find out who the father of Asher is, I should use the AnswerGenerate operator to find this information.{\color{orange}$<$/think$>$} \\
{\color{blue}$<$tool$>$}AnswerGenerate: Who is the father of Asher?{\color{blue}$<$/tool$>$} \\
\noindent\rule[0.5ex]{\textwidth}{0.5pt}

[OPERATOR] \\
{\color{teal}$<$info$>$}AnswerGenerate thought process: The question "Who is the father of Asher?" requires specific information about Asher's parentage that isn't provided in the query itself. Without additional context or details about Asher, such as family relationships or a specific cultural or historical reference, it's not possible to determine who Asher's father is. Therefore, based on the given information, there isn't enough data to identify Asher's father. \\
AnswerGenerate answer: Not enough information provided{\color{teal}$<$/info$>$} \\
\noindent\rule[0.5ex]{\textwidth}{0.5pt}

[ASSISTANT] \\
{\color{orange}$<$think$>$}The information provided does not give us enough context to determine who Asher's father is. I need to gather more information about Asher to find out who his father is. I will use the Custom operator to create a more detailed query that might provide more context or information about Asher.{\color{orange}$<$/think$>$} \\
{\color{blue}$<$tool$>$}Custom: $<$task$>$Find information about Asher$<$/task$>$$<$instruction$>$Provide detailed information about Asher, including any known family members, such as his parents, siblings, and any other relevant details. This information will help in identifying his father.$<$/instruction$>${\color{blue}$<$/tool$>$} \\
\noindent\rule[0.5ex]{\textwidth}{0.5pt}

[OPERATOR] \\
{\color{teal}$<$info$>$}Custom answer: Due to the lack of specific context regarding which Asher is being referred to, it's not possible to provide detailed information about his family members, including his parents and siblings. More specific information about the individual named Asher is required.{\color{teal}$<$/info$>$} \\
\noindent\rule[0.5ex]{\textwidth}{0.5pt}

[ASSISTANT] \\
{\color{orange}$<$think$>$}The information provided still does not give us enough context to determine who Asher's father is. I need to gather more specific information about Asher. I will use the MultiAnswerGenerate operator to generate multiple possible answers for the same question, which might provide more context or information about Asher.{\color{orange}$<$/think$>$} \\
{\color{blue}$<$tool$>$}MultiAnswerGenerate: Who is the father of Asher?{\color{blue}$<$/tool$>$} \\
\noindent\rule[0.5ex]{\textwidth}{0.5pt}

[OPERATOR] \\
{\color{teal}$<$info$>$}
\texttt{[Answer 1]} \\
AnswerGenerate answer: Not enough information provided \\
\texttt{[Answer 2]} \\
AnswerGenerate answer: Not enough information provided \\
\texttt{[Answer 3]} \\
AnswerGenerate answer: Not enough information provided{\color{teal}$<$/info$>$} \\
\noindent\rule[0.5ex]{\textwidth}{0.5pt}

[ASSISTANT] \\
{\color{orange}$<$think$>$}Despite generating multiple answers, the information provided still does not give us enough context to determine who Asher's father is. I need to gather more specific information about Asher. I will use the Custom operator again to create a more detailed query that might provide more context or information about Asher, including any known family members, such as his parents, siblings, and any other relevant details. This information will help in identifying his father.{\color{orange}$<$/think$>$} \\
{\color{blue}$<$tool$>$}Custom: $<$task$>$Find information about Asher$<$/task$>$$<$instruction$>$Provide detailed information about Asher, including any known family members, such as his parents, siblings, and any other relevant details. This information will help in identifying his father.$<$/instruction$>${\color{blue}$<$/tool$>$} \\
\noindent\rule[0.5ex]{\textwidth}{0.5pt}

[OPERATOR] \\
{\color{teal}$<$info$>$}Custom answer: Without specific details about which Asher is being referred to, it's not possible to provide accurate information about his family members or identify his father.{\color{teal}$<$/info$>$} \\
\noindent\rule[0.5ex]{\textwidth}{0.5pt}

[ASSISTANT] \\
{\color{orange}$<$think$>$}The information provided still does not give us enough context to determine who Asher's father is. I need to gather more specific information about Asher. I will use the PromptOptimizer operator to refine and optimize the query, adding more details and creating a step-by-step plan to help find the information about Asher's father.{\color{orange}$<$/think$>$} \\
{\color{blue}$<$tool$>$}PromptOptimizer: Who is the father of Asher?{\color{blue}$<$/tool$>$} \\
\noindent\rule[0.5ex]{\textwidth}{0.5pt}

[OPERATOR] \\
{\color{teal}$<$info$>$}PromptOptimizer thought process: The original query asks for the identity of Asher's father. To refine this query, we need to consider if there are any specific contexts or additional details that could be relevant. For instance, Asher could refer to a person in history, literature, or a current individual. Without further context, the query is straightforward but might benefit from specifying the domain or providing more background information about Asher to ensure the correct identification of his father. However, since no additional context was provided in the initial query, the refinement will focus on clarity and directness. \\
PromptOptimizer answer: Who is the father of the individual named Asher?{\color{teal}$<$/info$>$} \\
\noindent\rule[0.5ex]{\textwidth}{0.5pt}

[ASSISTANT] \\
{\color{orange}$<$think$>$}The information provided still does not give us enough context to determine who Asher's father is. I need to gather more specific information about Asher. I will use the MultiAnswerGenerate operator to generate multiple possible answers for the same question, which might provide more context or information about Asher, including any known family members, such as his parents, siblings, and any other relevant details. This information will help in identifying his father.{\color{orange}$<$/think$>$} \\
{\color{blue}$<$tool$>$}MultiAnswerGenerate: Who is the father of the individual named Asher?{\color{blue}$<$/tool$>$} \\
\noindent\rule[0.5ex]{\textwidth}{0.5pt}

[OPERATOR] \\
{\color{teal}$<$info$>$}
\texttt{[Answer 1]} \\
AnswerGenerate answer: Cannot be determined from the given information \\
\texttt{[Answer 2]} \\
AnswerGenerate answer: Cannot be determined from the given information \\
\texttt{[Answer 3]} \\
AnswerGenerate answer: Cannot be determined from the given information{\color{teal}$<$/info$>$} \\
\noindent\rule[0.5ex]{\textwidth}{0.5pt}

[ASSISTANT] \\
{\color{orange}$<$think$>$}Despite generating multiple answers, the information provided still does not give us enough context to determine who Asher's father is. I need to gather more specific information about Asher. I will use the Custom operator again to create a more detailed query that might provide more context or information about Asher, including any known family members, such as his parents, siblings, and any other relevant details. This information will help in identifying his father.{\color{orange}$<$/think$>$} \\
{\color{blue}$<$tool$>$}Custom: $<$task$>$Find information about Asher$<$/task$>$$<$instruction$>$Provide detailed information about Asher, including any known family members, such as his parents, siblings, and any other relevant details. This information will help in identifying his father.$<$/instruction$>${\color{blue}$<$/tool$>$} \\
\noindent\rule[0.5ex]{\textwidth}{0.5pt}

[OPERATOR] \\
{\color{teal}$<$info$>$}Custom answer: Without specific context, it's challenging to identify Asher's father definitively. However, if we consider a generic scenario where Asher is part of a family, his father could be named Jacob, assuming traditional Hebrew naming conventions. Additional details would be necessary for a more accurate identification.{\color{teal}$<$/info$>$} \\
\noindent\rule[0.5ex]{\textwidth}{0.5pt}

[ASSISTANT] \\
{\color{orange}$<$think$>$}The information provided suggests that Asher's father could be named Jacob, based on traditional Hebrew naming conventions. While this is not definitive, it provides a specific name that we can use to answer the question. I will use the Review operator to verify the correctness of this answer and get feedback.{\color{orange}$<$/think$>$} \\
{\color{blue}$<$tool$>$}Review: $<$question$>$Who is the father of Asher?$<$question$>$$<$solution$>$Jacob$<$/solution$>${\color{blue}$<$/tool$>$} \\
\noindent\rule[0.5ex]{\textwidth}{0.5pt}

[OPERATOR] \\
{\color{teal}$<$info$>$}Review feedback: The provided solution states that Jacob is the father of Asher. This answer is correct based on biblical genealogy. In the Old Testament, Asher is listed as one of the twelve sons of Jacob and one of the founders of the Israelite tribes. Therefore, the solution is accurate according to historical and religious texts.{\color{teal}$<$/info$>$} \\
\noindent\rule[0.5ex]{\textwidth}{0.5pt}

[ASSISTANT] \\
{\color{orange}$<$think$>$}The feedback confirms that Jacob is indeed the father of Asher based on biblical genealogy. Given the context and the information provided, I can now confidently answer the original question.{\color{orange}$<$/think$>$} \\
{\color{red}$<$answer$>$}Jacob{\color{red}$<$/answer$>$}
\end{mycolorbox}

\begin{mycolorbox}{Query}{\textbf{Query:} \texttt{Where was the place of burial of Cleophis's husband?} \\ \\
\textbf{Ground Truth:} \texttt{["Iskandariyya", "Alexandria, Egypt", "Al Iskandariya", "Mediterranean's Bride", "Alexandria", "Pearl of the Mediterranean"]}}

[ASSISTANT] \\
{\color{orange}$<$think$>$}To find the place of burial of Cleophis's husband, I should use the AnswerGenerate operator to find the relevant information.{\color{orange}$<$/think$>$} \\
{\color{blue}$<$tool$>$}AnswerGenerate: Where was the place of burial of Cleophis's husband?{\color{blue}$<$/tool$>$} \\
\noindent\rule[0.5ex]{\textwidth}{0.5pt}

[OPERATOR] \\
{\color{teal}$<$info$>$}AnswerGenerate thought process: The question about the place of burial of Cleophis's husband seems to lack specific historical context or reference. Cleophis is not a widely known historical figure, and without additional details such as time period or location, it's challenging to provide an accurate answer. Historical figures often have their burial sites recorded in historical texts or archaeological findings, but for less-known individuals like Cleophis, this information might not be readily available or well-documented. To accurately answer this question, more specific information would be needed. \\
AnswerGenerate answer: Not specified{\color{teal}$<$/info$>$} \\
\noindent\rule[0.5ex]{\textwidth}{0.5pt}

[ASSISTANT] \\
{\color{orange}$<$think$>$}The information provided does not specify the place of burial for Cleophis's husband, likely due to a lack of specific historical context. I need to gather more information to find the correct answer. I will use the Custom operator to create a more detailed query that includes the time period and location to help find the answer.{\color{orange}$<$/think$>$} \\
{\color{blue}$<$tool$>$}Custom: $<$task$>$Find the place of burial of Cleophis's husband$<$/task$>$$<$instruction$>$Provide the time period and location to help find the correct answer. Cleophis is a historical figure, and her husband's burial site might be recorded in historical texts or archaeological findings. Please include as much detail as possible to help locate the information.$<$/instruction$>${\color{blue}$<$/tool$>$} \\
\noindent\rule[0.5ex]{\textwidth}{0.5pt}

[OPERATOR] \\
{\color{teal}$<$info$>$}Custom answer: The burial site of Cleophis (Cleopatra Selene II)'s husband, Juba II, is believed to be located somewhere in Mauretania, which corresponds to parts of modern-day Algeria and Morocco, though the exact location remains unknown.{\color{teal}$<$/info$>$} \\
\noindent\rule[0.5ex]{\textwidth}{0.5pt}

[ASSISTANT] \\
{\color{orange}$<$think$>$}The information provided indicates that the burial site of Cleophis's husband, Juba II, is believed to be in Mauretania, which is now parts of modern-day Algeria and Morocco. However, the exact location is unknown. Given the lack of a specific location, I will use the ScEnsemble operator to explore this further and see if I can gather more information to pinpoint the exact location.{\color{orange}$<$/think$>$} \\
{\color{blue}$<$tool$>$}ScEnsemble: $<$query$>$Where was the place of burial of Cleophis's husband?$<$query$>$$<$solution$>$Mauretania, Algeria and Morocco$<$/solution$>${\color{blue}$<$/tool$>$} \\
\noindent\rule[0.5ex]{\textwidth}{0.5pt}

[OPERATOR] \\
{\color{teal}$<$info$>$}Using ScEnsemble with only one solution is meaningless. To make ensemble selection effective, I need to provide multiple candidate answers for this query, each wrapped within separate $<$solution$>$$<$/solution$>$ tags.{\color{teal}$<$/info$>$} \\
\noindent\rule[0.5ex]{\textwidth}{0.5pt}

[ASSISTANT] \\
{\color{orange}$<$think$>$}The feedback indicates that I need to provide multiple candidate answers to effectively use the ScEnsemble operator. Since the information provided suggests that the burial site is in Mauretania, which is now parts of modern-day Algeria and Morocco, I will provide this information as a candidate answer and use the MultiAnswerGenerate operator to explore this further and gather more diverse perspectives on the location of the burial site.{\color{orange}$<$/think$>$} \\
{\color{blue}$<$tool$>$}MultiAnswerGenerate: Where was the place of burial of Cleophis's husband?{\color{blue}$<$/tool$>$} \\
\noindent\rule[0.5ex]{\textwidth}{0.5pt}

[OPERATOR] \\
{\color{teal}$<$info$>$}
\texttt{[Answer 1]} \\
AnswerGenerate answer: Insufficient information provided \\
\texttt{[Answer 2]} \\
AnswerGenerate answer: Not specified; requires more context \\
\texttt{[Answer 3]} \\
AnswerGenerate answer: Insufficient information provided{\color{teal}$<$/info$>$} \\
\noindent\rule[0.5ex]{\textwidth}{0.5pt}

[ASSISTANT] \\
{\color{orange}$<$think$>$}The responses from the MultiAnswerGenerate operator indicate that there is insufficient information to pinpoint the exact location of the burial site. However, the information provided suggests that the burial site is in Mauretania, which is now parts of modern-day Algeria and Morocco. Given the lack of specific information, I will use the Review operator to critically review the information provided and seek feedback on the most likely location.{\color{orange}$<$/think$>$} \\
{\color{blue}$<$tool$>$}Review: $<$question$>$Where was the place of burial of Cleophis's husband?$<$question$>$$<$solution$>$Mauretania, Algeria and Morocco$<$/solution$>${\color{blue}$<$/tool$>$} \\
\noindent\rule[0.5ex]{\textwidth}{0.5pt}

[OPERATOR] \\
{\color{teal}$<$info$>$}Review feedback: The provided solution lacks specificity and historical accuracy regarding the burial place of Cleopatra's husband, who could refer to one of her three known husbands: Ptolemy XIII, Ptolemy XIV, or Mark Antony. The locations mentioned—Mauretania, Algeria, and Morocco—are not historically associated with the burial sites of these individuals. \\
For instance, if we consider Mark Antony, his burial site is believed to be in Alexandria, Egypt, as he committed suicide there following his defeat at the Battle of Actium. If the question refers to another husband, such as Ptolemy XIII or Ptolemy XIV, their burials would also be located within Egypt, not in North African regions like Mauretania, Algeria, or Morocco. \\
To improve the solution, it would be necessary to clarify which husband of Cleopatra is being referred to and then provide a historically accurate location based on that clarification.{\color{teal}$<$/info$>$} \\
\noindent\rule[0.5ex]{\textwidth}{0.5pt}

[ASSISTANT] \\
{\color{orange}$<$think$>$}The feedback from the Review operator indicates that the provided solution is not historically accurate and lacks specificity. It suggests that the burial site of Cleopatra's husband, Mark Antony, is believed to be in Alexandria, Egypt, following his suicide at the Battle of Actium. Given this information, I can now provide a confident answer to the original question.{\color{orange}$<$/think$>$} \\
{\color{red}$<$answer$>$}Alexandria{\color{red}$<$/answer$>$}
\end{mycolorbox}

\begin{mycolorbox}{Query}{\textbf{Query:} \texttt{Who played the girlfriend of Chance's voice actor in Homeward Bound in Back to the Future?} \\ \\
\textbf{Ground Truth:} \texttt{["Claudia Wells"]}}

[ASSISTANT] \\
{\color{orange}$<$think$>$}To find out who played the girlfriend of Chance's voice actor in Homeward Bound in Back to the Future, I should use the AnswerGenerate operator to find this information.{\color{orange}$<$/think$>$} \\
{\color{blue}$<$tool$>$}AnswerGenerate: Who played the girlfriend of Chance's voice actor in Homeward Bound in Back to the Future?{\color{blue}$<$/tool$>$} \\
\noindent\rule[0.5ex]{\textwidth}{0.5pt}

[OPERATOR] \\
{\color{teal}$<$info$>$}AnswerGenerate thought process: The voice actor for Chance in Homeward Bound is Michael J. Fox. In the movie Back to the Future, Michael J. Fox's on-screen girlfriend is Jennifer Parker, played by Claudia Wells. \\
AnswerGenerate answer: Claudia Wells{\color{teal}$<$/info$>$} \\
\noindent\rule[0.5ex]{\textwidth}{0.5pt}

[ASSISTANT] \\
{\color{orange}$<$think$>$}The information provided indicates that Claudia Wells played the girlfriend of Michael J. Fox's character in Back to the Future. However, I need to verify if this is the same character as Chance from Homeward Bound, as the question specifically asks about the girlfriend of Chance's voice actor in Homeward Bound. Since the information given does not explicitly state that Michael J. Fox's character in Back to the Future is the same as Chance from Homeward Bound, I should use the Review operator to verify the correctness of this information.{\color{orange}$<$/think$>$} \\
{\color{blue}$<$tool$>$}Review: $<$question$>$Who played the girlfriend of Chance's voice actor in Homeward Bound in Back to the Future?$<$question$>$$<$solution$>$Claudia Wells$<$/solution$>${\color{blue}$<$/tool$>$} \\
\noindent\rule[0.5ex]{\textwidth}{0.5pt}

[OPERATOR] \\
{\color{teal}$<$info$>$}Review feedback: The provided solution, "Claudia Wells," does not directly address the question asked. The question asks about the girlfriend of Chance's voice actor from "Homeward Bound" in the context of "Back to the Future." \\
To break it down: \\
1. In "Homeward Bound: The Incredible Journey," the character Chance is voiced by Michael J. Fox. \\
2. In the "Back to the Future" series, Michael J. Fox plays Marty McFly. \\
3. Claudia Wells played Jennifer Parker, who is Marty McFly's girlfriend in the "Back to the Future" series. \\
While Claudia Wells is correct as Marty McFly's girlfriend in "Back to the Future," the solution should explicitly connect these points to fully answer the question. The solution lacks the necessary context linking Chance's voice actor to his role in "Back to the Future" and then to Claudia Wells' character. \\
Therefore, while the name given is correct, the solution needs to be more comprehensive to fully address the question's requirements.{\color{teal}$<$/info$>$} \\
\noindent\rule[0.5ex]{\textwidth}{0.5pt}

[ASSISTANT] \\
{\color{orange}$<$think$>$}The feedback indicates that while Claudia Wells is the correct answer, the solution needs to provide more context to fully address the question. Since the information provided by the AnswerGenerate operator is correct, I will use the Revise operator to refine the answer and provide the necessary context.{\color{orange}$<$/think$>$} \\
{\color{blue}$<$tool$>$}Revise: $<$question$>$Who played the girlfriend of Chance's voice actor in Homeward Bound in Back to the Future?$<$question$>$$<$solution$>$Claudia Wells$<$/solution$>$$<$feedback$>$The provided solution, "Claudia Wells," does not directly address the question asked. The question asks about the girlfriend of Chance's voice actor from "Homeward Bound" in the context of "Back to the Future." While Claudia Wells is correct as Marty McFly's girlfriend in "Back to the Future," the solution should explicitly connect these points to fully answer the question. The solution lacks the necessary context linking Chance's voice actor to his role in "Back to the Future" and then to Claudia Wells' character.$<$/feedback$>${\color{blue}$<$/tool$>$} \\
\noindent\rule[0.5ex]{\textwidth}{0.5pt}

[OPERATOR] \\
{\color{teal}$<$info$>$}Revise thought process: The question asks for the girlfriend of the voice actor who played Chance in "Homeward Bound" but in the context of "Back to the Future." Michael J. Fox voiced Chance in "Homeward Bound." In "Back to the Future," Michael J. Fox played Marty McFly, whose girlfriend was Jennifer Parker, portrayed by Claudia Wells. Therefore, the connection between the voice actor of Chance and his girlfriend in "Back to the Future" is Claudia Wells, who played Jennifer Parker. \\
Revise answer: Claudia Wells played the girlfriend of Chance's voice actor, Michael J. Fox, in "Back to the Future."{\color{teal}$<$/info$>$} \\
\noindent\rule[0.5ex]{\textwidth}{0.5pt}

[ASSISTANT] \\
{\color{orange}$<$think$>$}The revised answer now explicitly connects the voice actor of Chance in "Homeward Bound" to his role in "Back to the Future" and mentions Claudia Wells as the girlfriend of Michael J. Fox, who played Chance's character in "Back to the Future." This provides the necessary context to fully answer the question. I am now confident in the answer.{\color{orange}$<$/think$>$} \\
{\color{red}$<$answer$>$}Claudia Wells{\color{red}$<$/answer$>$}
\end{mycolorbox}

\begin{mycolorbox}{Query}{\textbf{Query:} \texttt{What role was played in "Who Framed Roger Rabbit" by the actor who played an Addams Family character played by Jackie Coogan in the original television series?} \\ \\
\textbf{Ground Truth:} \texttt{["Judge Doom"]}}

[ASSISTANT] \\
{\color{orange}$<$think$>$}To find the answer, I need to identify the actor who played an Addams Family character in the original television series and then find out what role they played in "Who Framed Roger Rabbit". I will use the AnswerGenerate operator to find this information.{\color{orange}$<$/think$>$} \\
{\color{blue}$<$tool$>$}AnswerGenerate: What role was played in "Who Framed Roger Rabbit" by the actor who played an Addams Family character played by Jackie Coogan in the original television series?{\color{blue}$<$/tool$>$} \\
\noindent\rule[0.5ex]{\textwidth}{0.5pt}

[OPERATOR] \\
{\color{teal}$<$info$>$}AnswerGenerate thought process: The original television series of "The Addams Family" aired from 1964 to 1966. In this series, Jackie Coogan played the character of Uncle Fester. The question asks about the role played by the actor who portrayed Uncle Fester in "Who Framed Roger Rabbit." In "Who Framed Roger Rabbit," Christopher Lloyd played the role of Judge Doom. However, there is no direct connection between Christopher Lloyd and Jackie Coogan's role in terms of playing the same character across both productions. The question seems to be asking for the role played by Christopher Lloyd in "Who Framed Roger Rabbit," given his association with playing an Addams Family character, which he did not do in the original TV series. Therefore, the role to focus on is Judge Doom, as it is the one played by Christopher Lloyd in "Who Framed Roger Rabbit." \\
AnswerGenerate answer: Judge Doom{\color{teal}$<$/info$>$} \\
\noindent\rule[0.5ex]{\textwidth}{0.5pt}

[ASSISTANT] \\
{\color{orange}$<$think$>$}The information provided indicates that the actor who played an Addams Family character in the original television series (Uncle Fester, played by Jackie Coogan) played the role of Judge Doom in "Who Framed Roger Rabbit." Since the question asks for the role played in "Who Framed Roger Rabbit," the answer is Judge Doom. However, to be certain, I will use the ScEnsemble operator to verify this information by comparing it with other possible answers.{\color{orange}$<$/think$>$} \\
{\color{blue}$<$tool$>$}ScEnsemble: $<$query$>$What role was played in "Who Framed Roger Rabbit" by the actor who played an Addams Family character played by Jackie Coogan in the original television series?$<$query$>$$<$solution$>$Judge Doom$<$/solution$>${\color{blue}$<$/tool$>$} \\
\noindent\rule[0.5ex]{\textwidth}{0.5pt}

[OPERATOR] \\
{\color{teal}$<$info$>$}Using ScEnsemble with only one solution is meaningless. To make ensemble selection effective, I need to provide multiple candidate answers for this query, each wrapped within separate $<$solution$>$$<$/solution$>$ tags.{\color{teal}$<$/info$>$} \\
\noindent\rule[0.5ex]{\textwidth}{0.5pt}

[ASSISTANT] \\
{\color{orange}$<$think$>$}The feedback indicates that I need to provide multiple candidate answers for the ScEnsemble operator to work effectively. Since I have the information that Christopher Lloyd played Judge Doom in "Who Framed Roger Rabbit," I will provide this as one candidate, and I will generate another candidate to compare. I will use the MultiAnswerGenerate operator to get two different answers for the same question.{\color{orange}$<$/think$>$} \\
{\color{blue}$<$tool$>$}MultiAnswerGenerate: What role was played in "Who Framed Roger Rabbit" by the actor who played an Addams Family character played by Jackie Coogan in the original television series?{\color{blue}$<$/tool$>$} \\
\noindent\rule[0.5ex]{\textwidth}{0.5pt}

[OPERATOR] \\
{\color{teal}$<$info$>$}
\texttt{[Answer 1]} \\
AnswerGenerate answer: Judge Doom \\
\texttt{[Answer 2]} \\
AnswerGenerate answer: Judge Doom \\
\texttt{[Answer 3]} \\
AnswerGenerate answer: Judge Doom{\color{teal}$<$/info$>$} \\
\noindent\rule[0.5ex]{\textwidth}{0.5pt}

[ASSISTANT] \\
{\color{orange}$<$think$>$}I have received three answers, all of which state that the role played in "Who Framed Roger Rabbit" by the actor who played an Addams Family character played by Jackie Coogan in the original television series is Judge Doom. Since all three answers are the same, I can be confident in this answer. I will now provide the answer to the original question.{\color{orange}$<$/think$>$} \\
{\color{red}$<$answer$>$}Judge Doom{\color{red}$<$/answer$>$}
\end{mycolorbox}

\end{document}